\documentclass[journal]{IEEEtran}

\usepackage{cite}

\usepackage{xcolor}
\usepackage{multirow}
\usepackage{pbox}
\usepackage{amsmath}
\usepackage{amssymb}
\usepackage{hyperref}
\usepackage{url}
\usepackage{colortbl}
\usepackage{graphicx}
\usepackage{adjustbox}
\usepackage{subfig}
\usepackage{makecell}
\usepackage{float}
\usepackage{scalerel}
\usepackage{tikz}
\usepackage{float}
\usetikzlibrary{svg.path}

\definecolor{orcidlogocol}{HTML}{A6CE39}
\tikzset{
  orcidlogo/.pic={
    \fill[orcidlogocol] svg{M256,128c0,70.7-57.3,128-128,128C57.3,256,0,198.7,0,128C0,57.3,57.3,0,128,0C198.7,0,256,57.3,256,128z};
    \fill[white] svg{M86.3,186.2H70.9V79.1h15.4v48.4V186.2z}
                 svg{M108.9,79.1h41.6c39.6,0,57,28.3,57,53.6c0,27.5-21.5,53.6-56.8,53.6h-41.8V79.1z M124.3,172.4h24.5c34.9,0,42.9-26.5,42.9-39.7c0-21.5-13.7-39.7-43.7-39.7h-23.7V172.4z}
                 svg{M88.7,56.8c0,5.5-4.5,10.1-10.1,10.1c-5.6,0-10.1-4.6-10.1-10.1c0-5.6,4.5-10.1,10.1-10.1C84.2,46.7,88.7,51.3,88.7,56.8z};
  }
}

\newcommand\orcidicon[1]{\href{https://orcid.org/#1}{\mbox{\scalerel*{
\begin{tikzpicture}[yscale=-1,transform shape]
\pic{orcidlogo};
\end{tikzpicture}
}{|}}}}

\definecolor{lightgray}{rgb}{0.92, 0.92, 0.92}

\ifCLASSINFOpdf
\else
\fi
\begin{document}

\title{\vspace{-3mm} Rank Flow Embedding for Unsupervised and Semi-Supervised Manifold Learning}

\author{Lucas Pascotti Valem \orcidicon{0000-0002-3833-9072},
    ~Daniel Carlos Guimarães Pedronette \orcidicon{0000-0002-2867-4838} \\ 
    and~Longin Jan Latecki \orcidicon{0000-0002-5102-8244}. %\IEEEmembership{Senior Member,~IEEE \vspace{-5mm}}% <-this % stops a space
\thanks{The authors are grateful to Fulbright Commission, São Paulo Research Foundation - FAPESP (grants \#2018/15597-6 and \#2017/25908-6), Brazilian National Council for Scientific and Tech. Development - CNPq (grants \#309439/2020-5 and \#422667/2021-8), and Petrobras. This work was also partly supported by the National Science Foundation, USA Grant No. IIS-1814745.}% <-this % stops a space
\thanks{L. P. Valem and D. C. G. Pedronette are with the Department of Statistics,
Applied Math. and Computing, State University of São Paulo, Rio Claro, Brazil (e-mail: \{daniel.pedronette,lucas.valem\}@unesp.br).}% <-this % stops a space
\thanks{L. J. Latecki is with Department of Computer and Information Sciences, Temple University, Philadelphia, USA (e-mail: latecki@temple.edu).}% <-this % stops a space
}% <-this % stops a space
\markboth{Accepted version of paper published in IEEE Transactions on Image Processing}%
{Valem\MakeLowercase{\textit{et al.}}: Bare Demo of IEEEtran.cls for IEEE Journals}

\maketitle

\begin{abstract}
Impressive advances in acquisition and sharing technologies have made the growth of multimedia collections and their applications almost unlimited.
However, the opposite is true for the availability of labeled data, which is needed for supervised training, since such data is often expensive and time-consuming to obtain.
While there is a pressing need for the development of effective retrieval and classification methods, the difficulties faced by supervised approaches highlight the relevance of methods capable of operating with few or no labeled data.
In this work, we propose a novel manifold learning algorithm named Rank Flow Embedding (RFE) for unsupervised and semi-supervised scenarios.
The proposed method is based on ideas recently exploited by manifold learning approaches, which include hypergraphs, Cartesian products, and connected components.
The algorithm computes context-sensitive embeddings, which are refined following a rank-based processing flow, while complementary contextual information is incorporated.
The generated embeddings can be exploited for more effective unsupervised retrieval or semi-supervised classification based on Graph Convolutional Networks.
Experimental results were conducted on 10 different collections.
Various features were considered, including the ones obtained with recent Convolutional Neural Networks (CNN) and Vision Transformer (ViT) models.
High effective results demonstrate the effectiveness of the proposed method on different tasks: unsupervised image retrieval, semi-supervised classification, and person Re-ID.
The results demonstrate that RFE is competitive or superior to the state-of-the-art in diverse evaluated scenarios.
\end{abstract}

\begin{IEEEkeywords}
ranking, embedding, unsupervised, semi-supervised, manifold learning, person Re-ID
\end{IEEEkeywords}

% \fi
\IEEEpeerreviewmaketitle

%--------------------------------------------------
\vspace{-2mm}
\section{Introduction}
%--------------------------------------------------

\IEEEPARstart{C}{ontent}-based Image Retrieval (CBIR) is a central tool behind a diversified range of applications. In fact, it can be seen as technology that helps to organize digital picture archives by their visual content~\cite{PaperSurveyCBIR_Datta2008}, including a broad spectrum of approaches, from general object retrieval to medical diagnostics support and person re-identification~\cite{PaperSurveyCBIR_Datta2008,SurveyDeepRetr_2021,PaperCBIRSurvey_2017}.
A traditional task is given by a query-by-example arrangement, which consists in retrieving the most similar images to a query image defined by the user from an image collection~\cite{PaperDifByNN_BMVC18}.
While involving various challenges and the fundamental open problem of robust image understanding~\cite{PaperSurveyCBIR_Datta2008}, it can also be seen as a rank-centered task, once the retrieved images are expected to be ranked according to the user needs.

The ranking tasks performed by CBIR approaches typically rely on two basic steps: the image content representation itself and the similarity measurement of collection images to the query.
The image representation is concerned with mapping an image to a point in a high-dimensional feature space. 
The similarity measurement, in turn, relies on assessing how close representations of collection images are from the query point in the feature space~\cite{paperBFSTREE}.
Conventionally, it is accomplished by computing the pairwise dissimilarity between
feature representations in the Euclidean space~\cite{PaperRegDiff_PAMI19}.

Extensive advances have been made in image representation techniques over the last decades.
Originally, the extraction of global features defined the dominant approach, where a myriad of features were proposed, mainly based on visual properties such as shape, texture, and color.
The global features  gave rise to local feature strategies, based on Bag-of-Words (BoW) model, largely studied over a decade~\cite{PaperSIFTCNN_TPAMI2017}.
More recently, the success of deep neural networks on feature representation has made them a fundamental tool in image retrieval.
Models pre-trained on huge datasets are broadly used through transfer learning to extract features of images~\cite{PaperDecOnOff_AAAI19,SurveyDeepRetr_2021}.

Despite the huge advances in representation strategies, especially supported by recent deep features given by Convolutional Neural Networks (CNN) and Vision Transformers (ViT) models, a major limitation is associated with the pairwise formulation of similarity measurements.
In fact, both traditional and deep-based representations lie on manifolds in a high-dimensional space~\cite{PaperFSpectralRank_CVPR18} such that pairwise similarity measures are insufficient to reveal the intrinsic relationship between images. 
Instead, similarities can be estimated more accurately along the geodesic paths of the underlying data manifold~\cite{PaperRegDiff_PAMI19}. 
The goal of such strategies is to somehow mimic human behavior in judging the similarity among objects; i.e., by considering the context of other objects.

In this research direction, different approaches have been proposed to post-process pairwise measures in order to compute more global and effective similarity measures~\cite{PaperDifProc_CVPR2013,PaperDiffusionProcessCVPR_Latecki2009,paperBFSTREE,PaperRLSimPR2013,PaperTPG_PAMI2012,paperLHRR}.
Different techniques and a comprehensive terminology have been employed, all following the common objective of capturing the structural similarity information encoded in the datasets through unsupervised contextual analysis.
Such contextual-sensitive similarity measures have been successfully applied to capture the geometry of the underlying manifold in order to improve retrieval tasks.

Diffusion processes demonstrated high potential in capturing the underlying manifold structure~\cite{PaperRegDiff_PAMI19,PaperEffDifManifold_CVPR2017}. 
Diffusion processes use a weighted graph, where each image is represented by a node, and edge weights are defined by pairwise affinity values. The pairwise affinities are re-evaluated in the context of other images, by spreading the similarity values across the graph. Affinities are spread on the manifold, which in turn improves the retrieval scores~\cite{PaperDifProc_CVPR2013}. 
Several variants have been proposed~\cite{PaperDifProc_CVPR2013}, including methods capable of analyzing high-order similarity relationships~\cite{PaperRegDiff_PAMI19}. In addition, such approaches are supported by a strong mathematical background but are  often associated to high computational costs~\cite{paperRDPAC}.

Re-ranking and rank-based manifold learning methods constitute another representative category of unsupervised post-processing methods~\cite{PaperHelloNeig_CVPR2011,BeyondDP_BaiINS2015,PaperManLearReckNN_PR2017,PaperRLSimPR2013,paperBFSTREE}.
In fact, ranked lists provide a rich source of contextual information once they establish a similarity relationship among a set of images, in contrast to pairwise relations.  Additionally, the most relevant information in the ranked lists is located at top positions, which enables the development of efficient algorithms~\cite{PaperInfoSciences2014}.
Reciprocal similarity relationships~\cite{PaperManLearReckNN_PR2017,PaperNeigsVISAPP2014,PaperManifoldIVC2014} and rank correlation measures~\cite{PaperRL_PR2013,BeyondDP_BaiINS2015,PaperRankCor_ACMTOM2018} have been successfully applied by various approaches.

Graphs and embeddings are modeling tools that also have been demonstrating a high potential for contextual similarity analysis.
The shortest path in the graph is used to define the similarity between images in~\cite{PaperLearningPath_PR2011}.
Connected Components are exploited in ~\cite{paperCorGraph2016,PaperManLearReckNN_PR2017} for spreading confident similarity relationships.
Lately, hypergraphs have been exploited, mainly due to their capacity of representing high-order similarity information~\cite{paperLHRR,PaperRegDiff_PAMI19}.
More recently, approaches that learn a mapping function to an embedded space have been proposed that exhibit the capacity of generalizing to new data~\cite{PaperMinManifold_CVPR18}, but such approaches are still rarely considered in the literature.

On the other hand, unsupervised image retrieval and semi-supervised classification are well-known and largely studied tasks.  However, they remain challenging interconnected tasks, with many applications in diverse scenarios (person re-identification~\cite{Karanam2019Survey}, remote sensing~\cite{paperRemoteSensing}, medical imaging~\cite{paperAlzheimerCBIR}, and many others).
In spite of many advances, most of the approaches address one specific problem. 
Our contribution is an unsupervised rank-based approach capable of refining similarity information and computing a context-sensitive representation, which can be exploited for improving the effectiveness of both unsupervised retrieval and semi-supervised classification.
We propose a novel manifold learning algorithm named Rank Flow Embedding (RFE). The proposed method is based on different and complementary ideas recently exploited by manifold learning approaches in order to provide a better contextual representation of dataset objects. 
The algorithm computes rank-based embeddings which are refined along the processing flow for each step.
This approach constitutes a key innovation in the sense that constitutes an unsupervised contextual-sensitive method capable of computing a novel representation and not only a similarity measure.

Firstly, a rank-based formulation is used to define a hypergraph model capable of representing high-order similarity information encoded in ranked lists. 
The hypergraph is used for iterative re-ranking, based on the similarity among embeddings defined by hyperedges (\textit{h-embeddings}).
 Next, Cartesian product operations are performed on hyperedges for maximizing their similarity relationships.
 While hyperedges effectively represent regional relationships, broader similarity relationships are also relevant.
 In this direction,  hypergraph structures are also used to model a graph and define high-confident Connected Components (CCs), aiming at estimating class information of datasets.
The information encoded in the CCs is exploited for a new re-ranking step and  used as class representatives to compute low-dimensional embeddings.
Such embeddings, in turn, can be exploited for more effective semi-supervised classification tasks.

The proposed method presents various contributions and innovations regarding related work.
Among them:
\begin{itemize}
    \item Most unsupervised context-sensitive approaches establish a novel similarity measure~\cite{ReRankFewShot_NIPS21,paperBFSTREE,PaperRegDiff_PAMI19}, but not a novel representation. Beyond that, this work proposes a novel rank-based approach for learning context-sensitive representations. More effective representations are fundamental for many applications, including unsupervised retrieval and semi-supervised classification, scenarios in which the method was evaluated; 
    \item The proposed approach presents substantial innovations in the way of computing such representations. The embeddings and their encoded similarity information are refined through a flow of rank-based structures and operations. Although some strategies already have been individually  exploited (graphs~\cite{PaperLearningPath_PR2011}, hypergraphs~\cite{PaperRegDiff_PAMI19}, and connected components~\cite{paperCorGraph2016}), our work allows the sequential refinement of similarity information along these structures. In addition, the proposed approach includes relevant distinctions in how such structures are defined and used. More specifically:
        \emph{(i)} The hypergraph model used is defined based on a novel rank normalization function, proposed in this work and named as reciprocal sigmoid;
        \emph{(ii)} The computation of connected components is based on a ranking of candidate edges, which estimates the confidence of edges using the hypergraph embeddings. The strategy consists of  a novel approach proposed in this work;
        \emph{(iii)} The use of similarity to the connected components for defining the dimensions of novel representations is also an innovation proposed in this paper.
    \item The method can be used in scenarios where the queries are not part of the dataset (\textit{unseen queries}), which is fundamental for many real-world applications and has been little exploited by related work in post-processing methods. 
\end{itemize}

The effectiveness of the proposed method was confirmed with a wide and diversified experimental evaluation.
The experimental results were obtained on 10 public datasets, including traditional image retrieval benchmarks and person Re-ID datasets.
For each dataset, different features were considered including CNN and recent Vision Transformers features.
On semi-supervised classification, the evaluation considered the proposed RFE embedding classified by different Graph Convolutional Network (GCN) models.
An ablation study was also conducted in order to assess the impact of each step of the proposed method.
The experimental evaluation also considers comparisons with other state-of-the-art approaches on various datasets.
The results demonstrate the effectiveness of the proposed method on different tasks: unsupervised image retrieval, semi-supervised classification, and person Re-ID.

This paper is organized as follows.
Section~\ref{sec:formaldef} presents the formal definition of addressed problems.
Section~\ref{sec:rfe} presents the proposed RFE method.
Section~\ref{sec:exp_eval} describes the conducted experimental evaluation.
Finally, Section~\ref{sec:conclusion} states conclusions and discusses the possible future works.

%--------------------------------------------------
\section{Problem Formulation}
\label{sec:formaldef}
%--------------------------------------------------

This section discusses the notation used  and formal definitions of main tasks involved, mostly following related work~\cite{paperLHRR,paperBFSTREE,paperRDPAC}.
Each task is discussed in the following subsections.

%--------------------------------- Sub-Section -----------------------------------
\vspace{-3mm}
\subsection{Feature Extraction and Similarity Computing}
%---------------------------------------------------------------------------------

Although images are the focus of this paper, a more global definition using multimedia objects is used.  The content of multimedia objects is represented by a feature extraction procedure. A $d$-dimensional representation is obtained and allows the pairwise comparison between objects. 
The comparison can be computed by two functions defined as:

\begin{itemize}
\item $\epsilon$: $o_i$ $\rightarrow$ $\mathbb{R}^{d}$ is a function, which extracts a feature vector $v_{\hat{i}}$ from a multimedia object $o_i$; 
\item $\delta$: $\mathbb{R}^{d} \times \mathbb{R}^{d} \rightarrow \mathbb{R}^+$ is a function that computes the distance between two multimedia objects according to the distance between their corresponding feature vectors.
\end{itemize}
 
 The distance between two objects $o_{i}$, $o_{j}$ is computed as $\delta$($\epsilon(o_{i})$, $\epsilon(o_{j})$). 
 The Euclidean distance is often employed to compute $\delta$, although the proposed ranking method is independent of distance measures.
 A similarity measure $\rho(o_i,o_j)$ can be computed based on distance function $\delta$ and used for ranking tasks.
 The notation $\rho(i,j)$ is used along the paper.

%--------------------------------- Sub-Section -----------------------------------
\vspace{-3mm}
\subsection{Retrieval and Rank Model}
%---------------------------------------------------------------------------------

Let  $\mathcal{C}$=$\{o_{1},$ $o_{2},$ $\dots,o_{n}\}$ be a multimedia collection, where $n = |\mathcal{C}|$ denotes the size of the collection $\mathcal{C}$. The target task refers to retrieving multimedia objects (images, videos) from $|\mathcal{C}|$ based on their content.
Let $o_{q}$ denotes a query object. A ranked list $\tau_{q}$ can be computed in response to the query $o_{q}$ based on the similarity function $\rho$.
The top positions of ranked lists are  expected to contain the most similar objects to the query  object.

Since $\tau_{q}$ can be expensive to compute  when $n$ is high, the  ranked list considers only a sub-set of the collection.
Formally, let $\tau_q$ be a ranked list that contains only the $L$ most similar objects to $o_q$, where $L \ll n$.
Let $\mathcal{C}_L$ be a sub-set of the collection $\mathcal{C}$, such that $\mathcal{C}_L \subset \mathcal{C}$ and $|\mathcal{C}_L| = L$. 
The ranked list $\tau_q$ can be defined as a bijection from the set $\mathcal{C}_L$ onto the set $[L]=\{1,2,\dots,L\}$.
 For a permutation $\tau_q$, we interpret $\tau_q (i)$ as the position (or rank) of the object $o_i$ in the ranked list $\tau_q$. 
If $o_{i}$ is ranked before $o_{j}$ in the ranked list of $o_q$, i.e., $\tau_q(i) 
< \tau_q(j)$, then $\rho(q, i)$ $\geq$  $\rho(q, j)$.

Every object $o_{i} \in \mathcal{C}$ can be taken as a query $o_{q}$. A set of ranked lists $\mathcal{T}$ = $\{\tau_{1}, \tau_{2},$ $\dots,$  $\tau_{n}\}$ can also be obtained, with a ranked list for each object in the collection $\mathcal{C}$. 
Based on the rank model, the neighborhood set can also be defined.
Let $o_q$ be a multimedia object taken as query,  a neighborhood set $\mathcal{N}(q,k)$ that contains the $k$ most similar multimedia objects to $o_q$ can be defined as follows:

\vspace{-4mm}
\begin{equation}
\begin{split}
\mathcal{N}(q,k) = \{  \mathcal{S} \subseteq \mathcal{C}, |\mathcal{S}| = k \wedge \forall o_i \in \mathcal{S}, o_j \in \mathcal{C} - \mathcal{S} : \\ 
	 \tau_q(i) < \tau_q(j) \}.
\end{split}
\end{equation}

%--------------------------------- Sub-Section -----------------------------------
\vspace{-5mm}
\subsection{Rank-based Manifold and Representation Learning}
%---------------------------------------------------------------------------------

The proposed RFE method aims to capture the structure of the dataset manifold by exploiting the similarity information encoded in the set of ranked lists $\mathcal{T}$. As a result, the RFE are evaluated on two objectives: (\textit{i}) computing a more effective similarity measure and ranking result for unsupervised retrieval and; (\textit{ii}) computing a more effective embedding to represent each image, which can be used by other tasks, as semi-supervised classification.

Regarding unsupervised manifold learning, a new and more effective set of ranked $\mathcal{T}_r$ is computed with the aim of improving the effectiveness of ranking results. More formally, we can describe the method as function $f_m$:
\begin{equation}
\mathcal{T}_r = f_m(\mathcal{T})
\end{equation}

The aggregation problem is also considered, in which different sets of ranked lists $\{\mathcal{T}_1, \mathcal{T}_2, \dots, \mathcal{T}_d\}$ are taken as input aiming at computing a more effective set $\mathcal{T}_r$.

Regarding representation learning, the objective is to compute an embedding that provides a more effective representation for a given object $o_i$ based on the contextual similarity information encoded in $\mathcal{T}$. Formally, it can be defined as function $f_e$:
\begin{equation}
\mathbf{e}_i = f_e(\mathcal{T},o_i),
\end{equation}
where $\mathbf{e}_i$ is a vector on a $d_e$-dimensional embeding space.

%--------------------------------------------------
\section{Proposed Method}
\label{sec:rfe}
%--------------------------------------------------

\begin{figure*}
\vspace{-4mm}
    \centering
    \includegraphics[width=.99\textwidth]{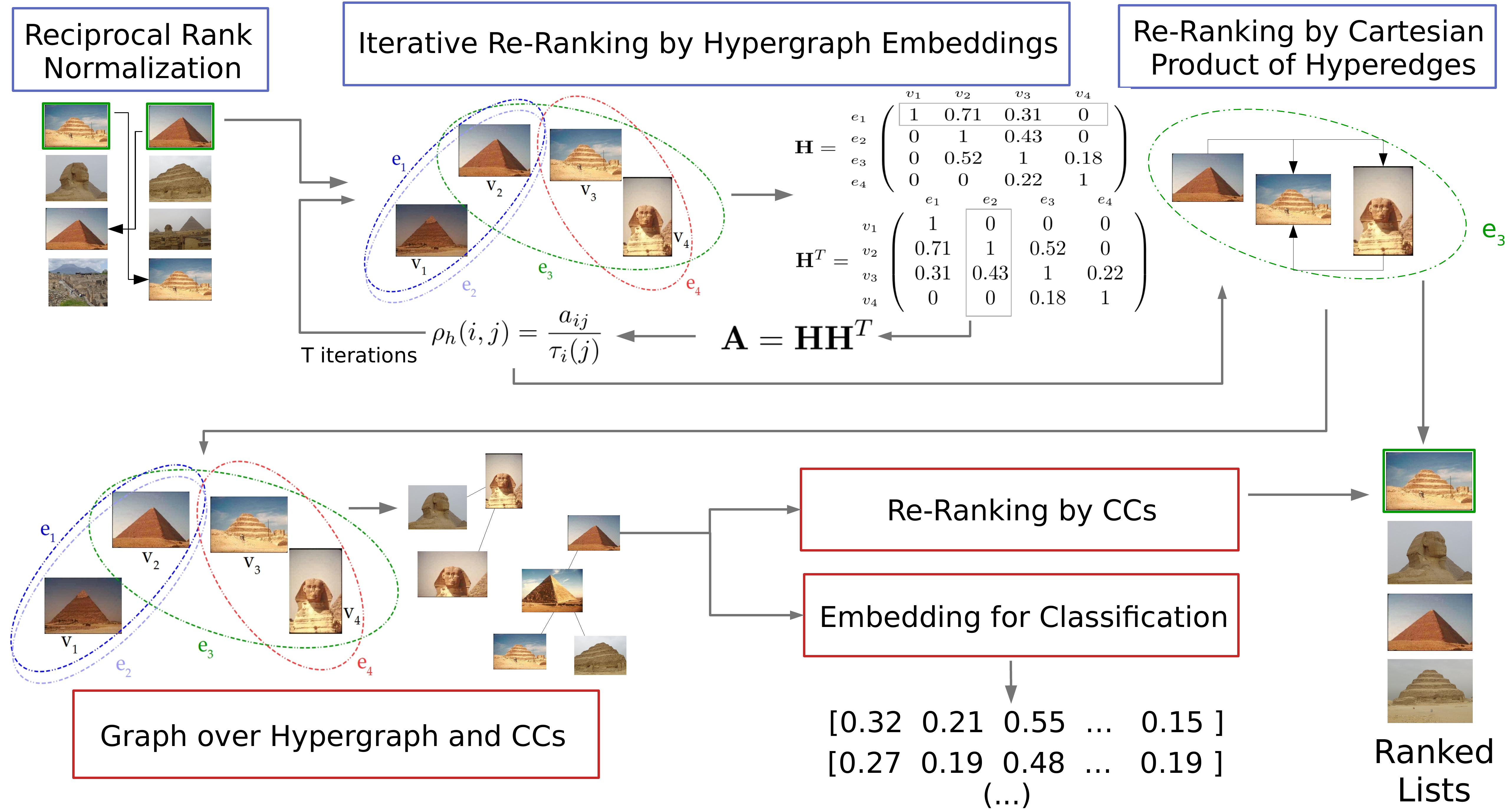}
    \caption{Overall organization of Rank Flow Embedding: in blue boxes the initial steps and in red boxes optional steps for refining retrieval and for computing embedding for semi-supervised classification.}
    \label{fig:rfe_diagram}
\vspace{-3mm}    
\end{figure*}

How to effectively design context-aware measures is a challenging question, which is closely associated with how to represent each image in terms of the collection in which is contained. 
Analogous to convolution and pooling operations used on CNNs, the proposed \textit{Rank Flow Embedding} (RFE) employ subsequent rank-based operations in order to define more effective contextual representations.
In fact, representations are derived from similarity to other images modeled by  rank information. Such representations, in turn, are used to derive more effective similarity measures. Such  mechanism is repeated through a flow of distinct and complementary operations in order to extract the maximum of available contextual information.

Figure~\ref{fig:rfe_diagram} presents the main steps of the proposed approach and the respective workflow. 
The proposed manifold learning algorithm can be used for unsupervised re-ranking, producing ranked lists as output retrieval results, or for representation learning, producing contextual vector representations.
The method can be summarized by the following steps:

\begin{enumerate}
    \item \textbf{Ranked Lists Normalization}: ranked lists are recomputed considering a sigmoid score computed based on the reciprocal ranked lists positions;
    \item \textbf{Re-ranking by Hypergraph Embeddings}: an iterative step that employs a hypergraph structure to analyze the underlying similarity information contained in the ranked lists. This step defines the h-embeddings and hyperedge weights, which are used by next steps;
    \item \textbf{Re-Ranking by Cartesian Product}: a  Cartesian  product  step  is  used  to  spread the  similarity information among elements in the same hyperedge;
    \item \textbf{Re-ranking by Connected Components}: high-confident connected components (CCs) are defined based on hypergraph structures (Step 2).   The CCs are computed based on the most confidential edges identified through the hyperedge weights. The CCs encode class information and cause objects in the same CC to have their similarities increased;
    \item \textbf{Embeddings by Connected Components}: more effective embeddings are computed for each dataset element considering their similarity to the identified CCs. This step is directed  for semi-supervised classification, since a low-dimensional embedding is obtained.
\end{enumerate}

Each stage is detailed and formally defined in the next sections.
In general, each step incrementally improves the effectiveness of rank-based similarity information and computes structures which are exploited in next steps.
While Steps 1-3 are suitable for general retrieval tasks, the Step 4 is  focused on datasets with larger similarity groups, in which information from CCs can be better exploited.
Hence Step 4 is not suitable for datasets with large numbers of very small classes.
Step 5 uses the constructed structures for computing embeddings used for classification.
Besides the standard retrieval pipeline and the embeddings for classification, rank aggregation tasks and the use of unseen queries are also discussed.

%..................................................
\vspace{-3mm}
\subsection{Rank Normalization by Reciprocal Sigmoid}
\label{subsec:rank_norm}
%..................................................

In opposite to the majority of distance measures, the ranking information is not symmetric.
The increase of symmetry generally produces a positive impact on the effectiveness of similarity information, widely exploited by reciprocal rank analysis~\cite{PaperManLearReckNN_PR2017,PaperHelloNeig_CVPR2011}.
However, most of the reciprocal approaches apply linear analysis to rank positions.
In this paper, we use a non-linear scoring function that assigns high weights to top-rank positions, with a fast decay around the neighborhood size, given by $k$. 
With this objective, a sigmoid function is applied.
Additionally, a higher relevance is assigned to the original rank position (squared) in comparison with the reciprocal rank position (linear).
The new similarity between objects $o_i$ and $o_j$ is defined by $\rho_{n}$:

\begin{equation} 
\label{EqRankNorm_rho}
	\rho_{n} (i,j) = \sigma (i,j)^2 \times \sigma(j,i).
\end{equation}

The function $\sigma$ which assigns weights according to rank positions is defined as:

\begin{equation} 
\label{EqRankNorm_sigma}
	\sigma (x,y) = 1 - \frac{1}{1+e^{-\alpha (\tau_{x}(y)-k/2))}},
\end{equation}

\noindent where $\alpha$ is a constant empirically evaluated in the experimental analysis.

Based on the measure $\rho_{n}$, which is computed between the objects in the top-$L$ positions, the ranked lists  are updated with a stable sorting algorithm.
The stable sorting is used in order to keep the position in the case of a tie.
An updated set of ranked list $\mathcal{T}_n$ is obtained as output.

%. . . . . . . .  . . . . . . . .  . . . . . . . .  
\vspace{-3mm}
\subsection{Re-Ranking by Hypergraph Embeddings}
%. . . . . . . .  . . . . . . . .  . . . . . . . .  

The contextual representation model used for data elements and how to exploit it to compute more effective similarity measures is a fundamental task in rank-based manifold learning.
In this work, we use a hypergraph model based on ranking information inspired by~\cite{paperLHRR,PaperHyperRankSSL_CVPR2010}.
The hypergraph establishes relations among set of objects, allowing to represent high-order similarity relationships.
The proposed RFE  method compute contextual embeddings based on hypergraph information and define an iterative re-ranking procedure based on comparison of such embeddings. 

%.......................................................
\subsubsection{\textbf{Hypergraph Embeddings}}
%.......................................................

Formally, a hypergraph model is defined by a tuple $H = (V, E_h, w)$, where $V$ represents a finite set of vertices and $E_{h}$ denotes the set of hyperedges.
The hyperedges set $E_h$ can be defined as the family of subsets of  $V$ such that $\bigcup_{e_i \in E_h} = V$.
A hyperedge $e_i$ is said to be incident to a vertex $v_j$ if $v_j \in e_i$.
For each hyperedge $e_i$, a positive weight $w(e_i)$ is assigned, which denotes the confidence of the relationships established by the hyperedge $e_i$.

Each vertex $v_i \in V$ represents an object in the collection: $o_i \in \mathcal{C}$.
For each object, a hyperedge is created by exploiting first and second-order neighborhood information.
A hyperedge $e_i$ is defined based on the neighborhood set of $o_i$ and its respective neighbors.
Formally, let  $o_x \in \mathcal{N}(i,k)$ be a neighbor of $o_i$ and let  $o_j \in \mathcal{N}(x,k)$ be a neighbor of $o_x$, the hyperedge $e_i$ is defined as:
\vspace{-3mm}
\begin{equation}
    \label{eq:ei_rfe}
    e_i =  \mathcal{N}(i,k) \bigcup_{o_x \in \mathcal{N}(i,k)}    \mathcal{N}(x,k) .
\end{equation}

Consequently, each image $o_i$
is now also represented by a hyperedge $e_i$.
Since the number of hyperedges is equal to the number of vertices, the obtained hypergraph can be represented by a square incidence matrix $\mathbf{H}_m$ of size $|E_h| \times |V|$, where elements $\mathbf{H}_m$ are define as:
\begin{equation}
	h_m (e_i, v_j) = 	\left\{ \begin{array}{ll}
		r(e_i,v_j), & ~\textrm{if } v_j \in e_i, \\
		0, & \textrm{ otherwise.}\\
 	\end{array} \right. 
\end{equation}

Row $i$ of $h_m$ tells which vertices belong to hyperedge $e_i$
and the score $r(e_i,v_j)$  indicates the degree of belonging of the vertex $v_j$ to hyperedge $e_i$.
The score $r$ is computed according to the number and relevance of mentions to $v_j$ in the hyperedge $e_i$ and is defined as:
\begin{equation}
\label{eqNr}
\begin{split}
	r(e_i,v_j) =  \sum_{o_x \in \mathcal{N}(i,k) \wedge o_j \in \mathcal{N}(x,k) }     w_p(i,x) \times w_p(x,j),
\end{split}  						 
\end{equation}
where $w_p(i,x)$ is a function that assigns a weight of relevance to $o_x$ according to the position in the ranked list $\tau_i$.
Notice that the score $r$ incorporates information from first and second-order ranking references, i.e., from neighbors and neighbors of neighbors.
The weight assigned to $o_x$ according to the position of the ranked list $\tau_i$ is defined by a log-based function as:
\begin{equation}
	w_p (i, x) = 1-\log_k \tau_i(x).
\end{equation}	

The function $w_p (i, x)$ reaches the maximum value of 1, which is assigned to the first position of the ranked lists and corresponds to the query image.
For the subsequent positions in the ranked lists, the function decays fast.

While the hyperedge $e_i$ provides a more comprehensive contextual representation for the object $o_i$, it can also be susceptible to noise in certain circumstances. As it considers second-order similarity relationships, non-relevant objects in rankings of neighbors can generate undesired references  in the hyperedge $e_i$.  With the aim of filter out such cases, we include a consistency check among hyperedges in order to obtain a more precise representation.

The main idea consists in verifying for each element in the hyperedge $e_i$ how it is referenced by other hyperedges. Most of objects in $e_i$ are expected to be relevant and compose a consistent set of  high-similarity among each other. Thus, a given relevant object $o_j \in e_i$ is expected to be referenced with high scores in the other hyperedges which represents most of elements in $e_i$. On the other hand, a noisy and non-relevant object $o_n \in e_i$ is not expected to be referenced in the same hyperedges.

In this way, the filtered score for a given object $o_j \in e_i$ is computed by multiplying scores in $e_i$ by the score of $o_j$ in hyperedges of elements referenced in $e_i$, which can be obtained by a matrix $\mathbf{H}$ computed as
\begin{equation} 
\mathbf{H} =  {\mathbf{H}_m}^2.
\end{equation}

The computation of matrix $\mathbf{H}$ defines the embeddings provided by the hypergraph model to represent each object, which we denote as \emph{h-embeddings}.
For an object $o_i$, its respective h-embedding can be defined by the correspondent row of matrix $\mathbf{H}$, such that:
\begin{equation} 
\mathbf{h}_i =  [h_{i1}, h_{i2}, \dots, h_{in}],
\end{equation}	
where $h_{ij}$ defines the similarity of object $o_j$ in the hyperedge $e_i$, also denoted as $h(i,j)$.

The definition of the hypergraph also includes a confidence of each hyperedge, given by the function $w(e_i)$.
A highly-effective hyperedge is expected to contain a consistent set of vertices. Therefore, it is expected to contain only a few vertices with high score values given by $h(e_i,\cdot)$.
Hence, the weight $w(e_i)$ is defined as:
\begin{equation}
	w (e_i) =  \sum_{j \in \mathcal{N}_h(i,k) }  h(i,j),
\end{equation}	
\noindent where $\mathcal{N}_h(i,k)$ is a neighborhood set defined among the elements with top $h(e_i,\cdot)$ score values in the hyperedge. 
The $\mathcal{N}_h$ set containing the vertices with the highest values of $h(e_i,\cdot)$ is formally defined as:
\begin{equation}
\begin{split}
	\mathcal{N}_h(q,k) = \{  \mathcal{S} \subseteq e_q, |\mathcal{S}| = k \wedge \forall o_i \in \mathcal{S}, o_j \in e_q - \mathcal{S}   : \\  h(q,i) > h(q,j) \}.
\end{split}	
\end{equation}	

Based on the previous equations, we can define a function $f_h(\cdot)$ that, given a set of ranked lists $\mathcal{T}_n$ as input, computes a hypergraph $H$ and its respective \emph{h-embeddings} given by the matrix $\mathbf{H}$. The function is defined as follows:
\begin{equation} 
(H,\mathbf{H}) = f_h (\mathcal{T}_n).
\end{equation}

In fact, the matrix $\mathbf{H}$  and the weight of edges $w(.)$ contain the main similarity information encoded in the hypergraph model. Both structures are exploited by the proposed RFE method and refereed along the paper.
Firstly, the information encoded in  matrix $\mathbf{H}$ is exploited to define a contextual similarity measure used for re-ranking.

%.......................................................
\subsubsection{\textbf{Hypergraph-based Re-Ranking}}
%.......................................................

While similar objects present similar ranked lists, it is expected that the respective h-embeddings are also similar.
Once the similarity information is encoded in the matrix $\mathbf{H}$, a similarity measure between two embeddings $\mathbf{h}_i$ and $\mathbf{h}_j$ can be computed by its product $\mathbf{h}_i\mathbf{h}_j$.
This operation can be modeled for all the objects by multiplying the matrix $\mathbf{H}$  by its transpose, with the objective of obtain the affinity matrix $\mathbf{A}$, defined as follows:
\begin{equation}
\mathbf{A} = \mathbf{H}   {\mathbf{H}}^T.
\end{equation}

The elements of matrix $\mathbf{A}$ given by $a_{ij}$ denote the similarity between objects $o_i$, $o_j$.
The matrix $\mathbf{A}$ contains most of the similarity information extracted based on the hypergraph, such that it  can be used to define a more effective similarity measure $\rho_h$.
In addition, the proposed measure also considers a residual similarity information, given by the original ranking position.
The measure is defined as:
\begin{equation}
\label{eq:rho_rfe}
\rho_h (i,j) = \frac{a_{ij}}{\tau_i(j)}.
\end{equation}

Based on the similarity computed by the function $\rho_h$, an updated set of ranked lists ${\mathcal{T}_h}^{(t)}$ is obtained by applying a stable sorting algorithm.
The ranked lists, in turn can be used to compute a novel hypergraph and the
 procedure can be iteratively repeated, such that the superscript $^{(t)}$ denotes the iteration.

After a certain number of $T$ iterations, the set of ranked lists ${\mathcal{T}_h}^{(T)}$ is provided to the function $f_h$, which returns a matrix $\mathbf{H}_{a}$ and a updated hypergraph $H_{a}$, used in next steps of the rank flow.
The index $a$ is used to indicate that they were obtained based on the affinity matrix:
\begin{equation} 
(H_{a},\mathbf{H}_{a}) = f_h ({\mathcal{T}_h}^{(T)}).
\end{equation}

%..................................................
\vspace{-3mm}
\subsection{Re-Ranking by Cartesian Product}
%..................................................

A Cartesian product step is used to expand the similarity information contained in the updated set of hyperedges $E_h^a$.
Inspired by~\cite{paperLHRR,PaperCPRR}, the procedure exploits high-order similarity relationships represented on hyperedges to compute more effective pairwise  measures.
Formally, given two hyperedges $e_q, e_i \in E_h^a$, the Cartesian product between them can be defined as:
\begin{equation}
 e_q \times e_i = \{ (v_x,v_y): ~ v_x \in e_q   \wedge   v_y \in e_i \}.
\end{equation}

The notation ${e_q}^2$ is used aiming to indicate the Cartesian product between elements of the same hyperedge $e_q$, such that $e_q \times e_q= {e_q}^2$.
For each pair of vertices $(v_i,v_j) \in {e_q}^2$ a pairwise relationship $p: E_h^a \times V \times V \rightarrow  \mathbb{R}^+$ is established.

A value $p$ is computed based on the weight $w(e_q)$, which indicates the level of confidence of the hyperedge that originated the association.
As previously mentioned, the weight $w (e_i)$ can  be interpreted as the confidence estimations of associations encoded on hyperedge $e_i$ 
The degrees of association of $v_i$ and $v_j$ are defined by:
\begin{equation}
p (e_q,v_i,v_j) = w(e_q) \times h(e_q,v_i) \times h(e_q,v_j).
\end{equation}

A pairwise similarity measure based on the Cartesian product is defined considering relationships contained in all the hyperedges.
This formulation presents the idea of exploiting the co-occurrence of $v_i$ and $v_j$ in different hyperedges, performing a sum of all the values of $p(\cdot,v_i,v_j)$:
\begin{equation}
\rho_c (i,j) = \sum_{e_q \in E \wedge  (v_i,v_j) \in {e_q}^2 }  \;   p (e_q,v_i,v_j).
\end{equation}

Based on the similarity function $\rho_c$, a more effective set of ranked lists $\mathcal{T}_c$ is computed by a stable sorting algorithm.
The ranked lists set $\mathcal{T}_c$ is provided to the function $f_h$ that computes an updated hypergraph and h-embeddings.
The index $c$ is used to indicate that they were obtained after the Cartesian product step:
\begin{equation} 
(H_{c},\mathbf{H}_{c}) = f_h ({\mathcal{T}_c}).
\end{equation}	

%..................................................
\vspace{-3mm}
\subsection{Graph over Hypergraph and Connected Components}
%..................................................
\label{sec:def_rerank_cc}

Although the hypergraph model provides an effective tool to represent regional similarity information, it does not represent the similarity among objects in the same class/cluster but more distant in the dataset manifold.
In order to represent such information, a high-confident graph is defined based on h-embeddings computed after Cartesian product operations.
The Connect Components are extracted from this graph and are used to represent class information and the global structure of similarity relationships encoded in the dataset.

%..................................................
\subsubsection{\textbf{Graph Definition}}
%..................................................

Formally, the graph is defined as $G$ = $(V, E)$, such that the set of vertices $V = \mathcal{C}$, where each node represents a collection object.
The set of edges $E$ is computed based on information provided by the hypergraph representation.
Firstly, a set of candidate edges $\mathcal{E}_c$ is defined based on the neighborhood set of each object as:
\begin{equation}
\mathcal{E}_c = \bigcup_{q \in V} \bigcup_{i \in \mathcal{N}(q,k)} \{ (q,i) \}. 
\end{equation}

In order to select the most confident edges, the set of candidates are ranked.
The ranked list $\tau_{c}$ is defined as a permutation of the set of candidate edges $\mathcal{E}_c$.
The permutation $\tau_c$ is the bijection of the set $\mathcal{E}_c$ onto the set $[n_k]=\{1,2,\dots,n_k\}$, 
The position of the pair $(q,i)$ in the ranked list is denoted by $\tau_c ((q,i))$.
The permutation is defined such that if $(q,i)$ is ranked before $(j,l)$, e.g, $\tau_c((q,i)) < \tau_c((j,l))$, then  $s_c(q, i)$ $\geq$  $s_c(j, l)$.
The function  $s_c$ is a similarity measure attributed to pairs based on the similarity between h-embeddings and confidence of the hyperedge, defined as:
\begin{equation}
	s_c(i,j)  =  \mathbf{h}_{c_{i}} {\mathbf{h}_{c_j}^{{T}}} \times w(e_i) \times w(e_j),
\end{equation}
\noindent where the pair $(i,j)$ identifies a pairs of hyperedges $e_i, e_j \in E_h^c$, and $E_h^c$ denotes a set of hyperedges of the hypergraph $H_c$.
Once ranked, a threshold should be established in order to defined the number of edges that are created. The threshold $t_c$ is defined as: 
\begin{equation}
	t_c = \frac{\sum_{e_q \in E_h^c} w(e_q)}{2 \times n}.
\end{equation}

The edge set $E$ is be defined using the threshold $t_c$ as
\begin{equation}
	E = \{(o_q, o_i) \mid (q,i) \in \mathcal{E}_c  \wedge  \tau_c((q,i) < t_c \}.
\end{equation}

The process of building the graph can be understood as a function $f_g$ that receives as input a hypergraph $H_c$ and a matrix
$\mathbf{H}_c$ (output of the Cartesian product) and computes a graph $G$:
\begin{equation} 
G = f_g (H_c,\mathbf{H}_c).
\end{equation}

%..................................................
\subsubsection{\textbf{Connected Components}}
%..................................................

Based on the defined graph, its respective Connected Components (CC) are extracted.
Formally, each CC is defined as a set of objects $\mathcal{C}_i$.
Given two objects $o_i$, $o_j$ $\in$ $\mathcal{C}_l$, there is a path (edge) between $o_i$, $o_j$.
Search algorithms in graphs (e.g. Depth and Breadth-First) and Tarjan algorithm can be used to compute the CCs.
The output for the dataset is provided by the set of connected components
$\mathcal{S} =$ $\{ \mathcal{C}_1, \mathcal{C}_2, \dots , \mathcal{C}_m \}$, such that $\bigcup_{\mathcal{C}_i \in \mathcal{S}} =  \mathcal{S}$ and $\bigcap_{\mathcal{C}_i \in \mathcal{S}} = \emptyset $.

The connected components are sets of similar objects and it is expected that such structures encode the information of sets or classes of the dataset.
Following this reasoning, an embedding is created based on the  \textit{h-embeddings} of the elements that are part of it.
Given a connected component $q$, the cc-embedding $\mathbf{c}_q$ is defined as:
\begin{equation} 
\mathbf{c}_q = \sum_{o_i \in \mathcal{C}_q} \mathbf{h}_{c_{i}}.
\end{equation}

Once the Connected Components (CCs) encode information associated to representation of classes, the similarity to such CCs embeddings can be exploited for computing a more globally contextual similarity measure.
In this way, a novel embedding  is computed for each object according to its similarity to the CCs embeddings.
Formally, let $\mathbf{e}_q$ be an embedding of an object of index $q$.
The computation of the value of position $i$ of this vector (embedding) is done as follows:
\begin{equation} 
\mathbf{e}_q[i] = \mathbf{h}_{c_q} \mathbf{c}_i^T,
\end{equation}	
where $i$ identifies the connected component $\mathcal{C}_i \in \mathcal{S}$ and $\mathbf{c}_i$ denotes the embedding that corresponds to this CC.
In this way, the embeddings can be computed for each element of the dataset.

%..................................................
\subsubsection{\textbf{Re-Ranking by Connected Components}}
%..................................................

The re-ranking by CCs exploits information about elements in the same CC.
In this way, the elements that present high similarity values in the same CC, have their similarities increased.
The first step of this process consists into define the $k$ elements with the highest values in each connected component.
A neighborhood set $\mathcal{N}_c(q,k)$ is defined for each element of index  $q$ considering a constant $k$:
\begin{equation}
\begin{split}
\mathcal{N}_c(q,k) = \{  \mathcal{S} \subseteq \mathcal{C}, |\mathcal{S}| = k \wedge \forall o_i \in \mathcal{S}, o_j \in \mathcal{C} - \mathcal{S} : \\ 
	\mathbf{c}_q[i] > \mathbf{c}_q[j] \}.
\end{split}
\end{equation}

The ranked list $\tau_{c_q}$ can be defined as the permutation of objects that have the $k$ highest values in the embedding $\mathbf{c}_q$.
The permutation is defined as the bijection of the set $\mathcal{N}_c(q,k)$  to the set $[k]=\{1,2,\dots,k\}$.
The position of an object $o_i$ in the ranked list computed by the embedding of the connect component  $\mathbf{c}_q$ is defined as  $\tau_{c_q} (i)$.
If $o_i$  is ranked before $o_j$ in a ranked list, this means, $\tau_{c_q}(q,i) < \tau_{c_q}(q,j)$, therefore $\mathbf{c}_q[i] \geq \mathbf{c}_q[j]$.

The re-ranking by CCs exploits three complementary information: (\textit{i}) the similarity between embeddings; (\textit{ii}) the object belonging to the same connected component and; (\textit{iii}) the residual information of rank position.
The similarity $\rho_e (i,j)$ is defined in order to combine such information, formally defined as:
\begin{equation}
    \rho_e (i,j) =  \sum_{o_i,o_j \in \mathcal{N}_c(q,k) } \frac{1+ \sqrt{\tau_{c_q}(q,i)^2 + \tau_{c_q}(q,j)^2} \times \mathbf{e}_i \mathbf{e}_j^T}{\tau_i(j)}.
\end{equation}

Based on the similarity function $\rho_e$, a set of ranked lists ${\mathcal{T}_e}$ is obtained by a stable sorting algorithm.
The set of ranked lists $\mathcal{T}_e$ is provided to the function  $f_h$ that computes a new hypergraph $H$ and matrix $\mathbf{H}$.
The index $e$ is used to indicate that they were obtained after the step of the connected components:
\begin{equation} 
(H_e,\mathbf{H}_e ) = f_h ({\mathcal{T}_e}).
\end{equation}	

%..................................................
\vspace{-3mm}
\subsection{Embeddings for Classification}
%..................................................

The class information encoded in the re-ranking by CCs can be useful for other machine learning tasks.
In this way, novel representations  are computed for dataset objects and used as embeedings for semi-supervised classifiers.
Given the ranked lists ${\mathcal{T}_e}$ and the hypergraph $H_e$ obtained in the previous step, we  obtain a graph with the updated connected components following the same equations defined in Section~\ref{sec:def_rerank_cc}.
Thus, the updated graph is defined as follows:
\begin{equation} 
G_e = f_g (H_e,\mathbf{H}_e).
\end{equation}	

The new connected components, considering the component $\mathbf{c}$ after the step of CC (index $e$) for the element $q$, are obtained as follows:
\begin{equation} 
\mathbf{c}_{e_q} = \sum_{o_i \in \mathcal{C}_{e_q}} \mathbf{h}_{e_i} .
\end{equation}	

Finally, each of the positions of the embedding vector, which is going to be used for classification, computed as follows:
\begin{equation} 
\mathbf{e}_{e_q}[i] = \mathbf{h}_{e_q} \mathbf{c}^{T}_{e_i},
\end{equation}	

\noindent where the index $e$ indicates that the variables were obtained after the re-ranking by the connect components.
The contextual embedding $\mathbf{e}_{e_q}$ is used as features by semi-supervised classifiers.

%..................................................
\subsection{Unseen Queries}
%..................................................

The formulation used by RFE considered an already known dataset, where all the elements of the dataset can be taken as queries.
However, RFE also allows to perform queries with elements that does not belong to the dataset, in a formulation known in the literature as  unseen queries.
To make this possible, RFE follows a strategy proposed in~\cite{PaperDecOnOff_AAAI19} by decoupling off-line procedures (for the whole dataset) of on-line procedures (for the unseen query).

On off-line setting, the conventional steps of the method (normalization, re-ranking by embeddings, Cartesian product, re-ranking by connected components) are normally executed for all the known elements in the dataset.
So, when  a new external query (unseen query) need to be evaluated, the $k$ most similar elements are computed for each of them and a h-embedding is generated for the new query.
The cosine distance between the query embedding and pre-computed embeddings in the  whole dataset is used to rank the unseen query, producing the ranked lists for such elements.

\subsection{Rank Aggregation}

The RFE can also be exploited to fuse different features, in rank aggregation tasks.
Different ranked lists sets $\{\mathcal{T}_1, \mathcal{T}_2, \dots, \mathcal{T}_d\}$ are used as input with the objective of computing a more effective output set $\mathcal{T}_r$.
The normalization step is performed individually for each of the rankers and the values are accumulated in a single sparse matrix $M_{f}$, once only top-$L$ positions are considered.
New ranked lists ${\mathcal{T}_{f}}$ are obtained by the sorting objects based on scores given by the matrix $M_{f}$.
After that, the RFE (which can be understood as a function $f_r$) is executed for the ranked lists ${\mathcal{T}_{f}}$ and the 
list  $\mathcal{T}_r$ is obtained as result:
\begin{equation}
\mathcal{T}_r = f_m(\mathcal{T}_{f}).
\end{equation}

%--------------------------------------------------
\section{Experimental Evaluation}
\label{sec:exp_eval}
%--------------------------------------------------

This section discusses the experimental evaluation conducted to assess the effectiveness of the proposed method.
Section~\ref{sec:ExpSet} describes the datasets and experimental settings.
Section~\ref{sec:param} discusses the impact of parameters while Section~\ref{sec:Ablation} presents an ablation study that includes an analysis of the impact of each step in our proposed method.
Section~\ref{sec:RetResults} and~\ref{sec:clasifResults} present the results on unsupervised retrieval and semi-supervised classification tasks, respectively.
The results for unseen queries are described in Section~\ref{sec:unseenQueries}.
Sections~\ref{sec:comparisonRetrieval} and~\ref{sec:comparisonClassification} compare RFE with other state-of-the-art approaches for retrieval and classification, respectively.
Finally, Section~\ref{sec:visual} presents a visual analysis for both tasks.

%....................................................
\subsection{Experimental Settings}
\label{sec:ExpSet}
%....................................................

A broad experimental evaluation was conducted on 10 different image datasets, which are presented in Table~\ref{tab:datasets}. The datasets vary in size from 400 to 72,000 images. 
In this work, there are two different experimental scenarios: \emph{(i)} unsupervised image retrieval, which was assessed on all datasets; and \emph{(ii)} semi-supervised image classification conducted on the Flowers and Corel5k datasets.
The retrieval category encompasses not only general-purpose image datasets, but also person Re-ID datasets (i.e., CUHK03, Market, Duke).

Due to the highly diverse aspects of each dataset, we employed different evaluation measures in each case to enable comparisons with other approaches.
In the classification task, we used accuracy as the evaluation measure. In contrast, for the retrieval task, other measures were used, with Mean Average Precision (MAP) being the most common.
For Re-ID datasets, the R1 (which, in this case, is equivalent to Precision@1) was included, since it is commonly reported in the literature. For the UKbench dataset, which has the smallest number of images per class (only 4), the NS-Score was used.
The NS-Score is the average of correct images at the top-4 positions of the ranked lists.

We adopted the evaluation protocol for each dataset based on common practices in the literature.
For most of them, all the images were considered as queries, except for Holidays~\cite{JegouECCV2008} and Re-ID ones, where a different protocol was adopted~\cite{cuhk03_new_protocol, market1501, dukemtmc}.
For Holidays, there is a specific set of queries~\cite{JegouECCV2008}.
Regarding Re-ID, each dataset has a set of queries and a corresponding gallery set~\cite{cuhk03_new_protocol, market1501, dukemtmc}, which is the set of images that are ranked in relation to the query.

\begin{table}[ht!]
\caption{Datasets considered in the experimental evaluation.}
\label{tab:datasets}
\centering
\resizebox{.47\textwidth}{!}{ 
\begin{tabular}{l|c|c|c}
\hline
\textbf{Dataset Name}    & \textbf{Num. of} & \textbf{Dataset} & \textbf{Evaluation} \\ 
\textbf{}    & \textbf{Classes} & \textbf{Size} & \textbf{Measures} \\ \hline
\textbf{ORL Faces~\cite{paperORLFaces}}   & 40           & 400                       & Recall@15          \\ %\hline
\textbf{Flowers~\cite{PaperFlowers}}   & 17           & 1,360                       & Accuracy, MAP          \\ %\hline
\textbf{MPEG-7~\cite{PaperMPEG7_Latecki2000}}   & 70           & 1,400                       & Recall@40          \\ 
%\hline
\textbf{Holidays~\cite{JegouECCV2008}}   & 500           & 1,491                       & MAP          \\ %\hline
\textbf{Corel5k~\cite{PaperCorel5k_PR2013}}   & 50           & 5,000                       & Accuracy, MAP          \\ %\hline
\textbf{UKBench~\cite{PaperUKBench_CVPR2006}}   &   2,550         & 10,200                       & NS-Score, MAP          \\ %\hline
\textbf{CUHK03~\cite{cuhk03, cuhk03_new_protocol}}     & 1,467           & 14,097                          & R1, MAP~\cite{cuhk03_new_protocol}          \\ %\hline
\textbf{Market1501~\cite{market1501}} & 1,501           & 32,217                           & R1, MAP~\cite{market1501}          \\ %\hline
\textbf{DukeMTMC~\cite{dukemtmc}}   & 1,812           & 36,411                       & R1, MAP~\cite{dukemtmc}          \\ 
\textbf{ALOI~\cite{IJCV_2005_Geusebroek}}   & 1,000           & 72,000                       & MAP          \\ %\hline
\hline
\end{tabular}
}
\end{table}

A comprehensive set of descriptors (features) were used considering both traditional and deep learning extractors, including Convolutional Neural Networks (CNNs) and Visual Transformers (VIT).
For most of the datasets, a similar set of descriptors were used to keep the evaluation consistent.
All the CNNs were trained on ImageNet dataset\footnote{\url{https://github.com/Cadene/pretrained-models.pytorch}}.
For Re-ID datasets (i.e., CUHK03, Market, Duke), we used CNNs which are more specific for Re-ID and trained on the MSMT17 dataset, extracted using \emph{torchreid}\footnote{\url{https://github.com/KaiyangZhou/deep-person-reid}}.

The semi-supervised classification relies on Graph Convolutional Networks (GCNs), which are stochastic.
Since the results of the executions vary, we report an average of 5 executions on 10 different folds.
This was adopted for our method and all the baselines.
For unsupervised retrieval, the executions are deterministic.

%....................................................
\subsection{Parametric Space Analysis}
\label{sec:param}
%....................................................

\begin{figure}[!ht]
    \centering
    \includegraphics[width=.37\textwidth]{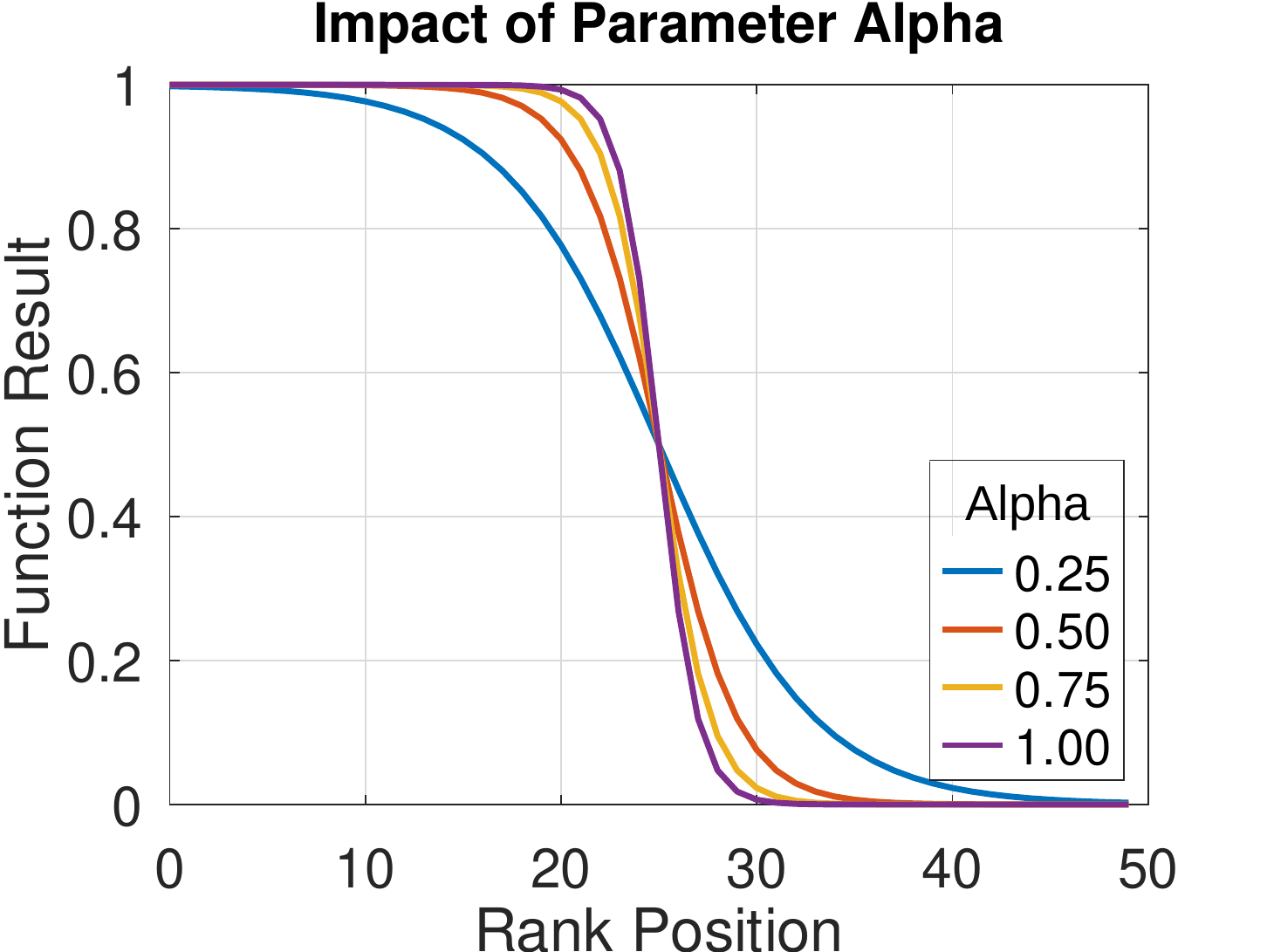}
    \caption{Impact of parameter $\alpha$ in function $\sigma$ (Equation~\ref{EqRankNorm_sigma}) as the rank position varies.}
    \label{fig:alpha}
\end{figure}

Initially, an experiment was conducted to visualize the impact of parameter $\alpha$ in the reciprocal sigmoid function, which is used in order to compute the rank normalization.
This is the first step of our proposed approach, described in Section~\ref{subsec:rank_norm}.
The normalization mainly relies on Equation~\ref{EqRankNorm_sigma}, which defines a reciprocal sigmoid function ($\sigma$).
Figure~\ref{fig:alpha} presents the values for Equation~\ref{EqRankNorm_sigma} ($\sigma$ in y-axis) as the Rank Position ($\tau_{x}(y)$ in x-axis) varies.
Different values of alpha were considered.
The figure reveals that $\alpha$ is responsible for changing the steepness of the sigmoid curve, which refers to how quickly the output of the function changes as the input (i.e., the rank position) increases.
However, it is challenging to determine an appropriate value of $\alpha$ based solely on this plot.

\begin{figure*}[th!]
\centering
\subfloat[Flowers]{\includegraphics[width=.34\textwidth]{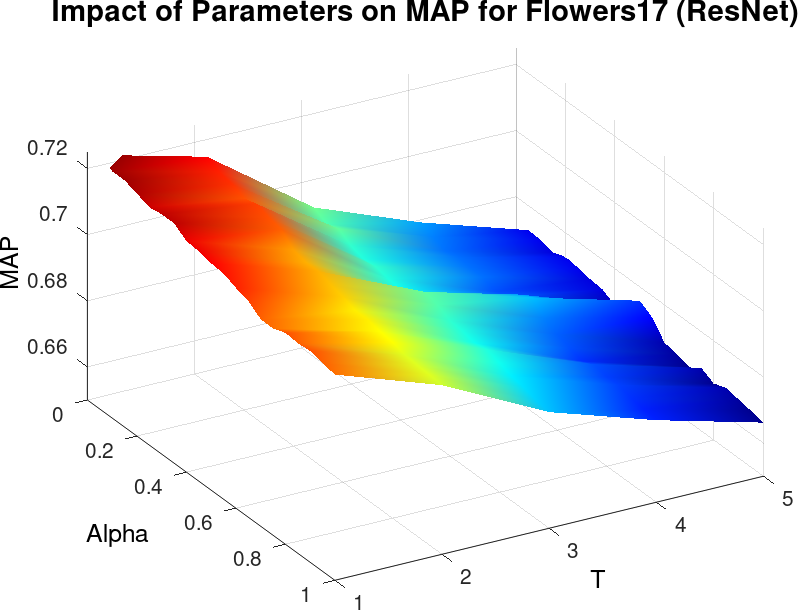}%
\label{flowers_param}}
\hspace{0.20cm}
\subfloat[Corel5k] {\includegraphics[width=.34\textwidth]{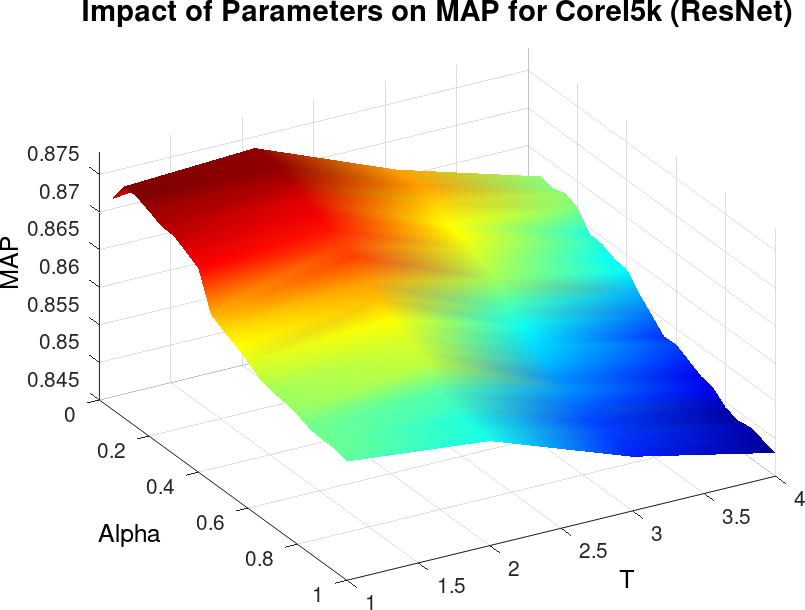}%
\label{corel_params}}
\caption{Impact of parameters $\alpha$ and T (number of iterations) on MAP for two datasets.}
\label{fig:param_surfaces}
\end{figure*}

Based on this issue, an analysis was conducted with the objective of identifying default parameters.
Figure~\ref{fig:param_surfaces} presents the impact of parameters $\alpha$ and T (number of iterations) on the MAP results for two datasets (i.e., Flowers and Corel5k).
The CNN-ResNet~\cite{paperRESNET} was considered for this experiment.
Since we are not evaluating the parameter $k$ in this case, we set it to the number of elements per class ($k=100$).
This is done to keep the focus of the analysis on $\alpha$ and T.
The surface shows that the lowest values of $\alpha$ and $T$ are more appropriate.
Notice that the set of parameters ($T$, $\alpha$) = $(2, 0.1)$ is close to the best results in all cases (a, b, and c).
Therefore, we used these values for all subsequent experiments.

%....................................................
\subsection{Ablation Study}
\label{sec:Ablation}
%....................................................

An ablation study was conducted to analyze the effectiveness of each step of the proposed method on 6 different datasets.
We evaluated the retrieval results incrementally from Steps 1 to 4, as discussed in Section~\ref{sec:rfe}.
Step (0) corresponds to the original features, Step (1) involves ranked lists normalization, Step (2) performs re-ranking by hypergraph embeddings, Step (3) computes re-ranking by Cartesian product, and Step (4) re-ranks by connected components. In this case, we excluded Step (5), which generates embeddings, as it is only necessary for semi-supervised classification.

\begin{figure*}[bth!]
\vspace{-2mm}
\centering
\subfloat[Flowers]
{\includegraphics[width=.34\textwidth]{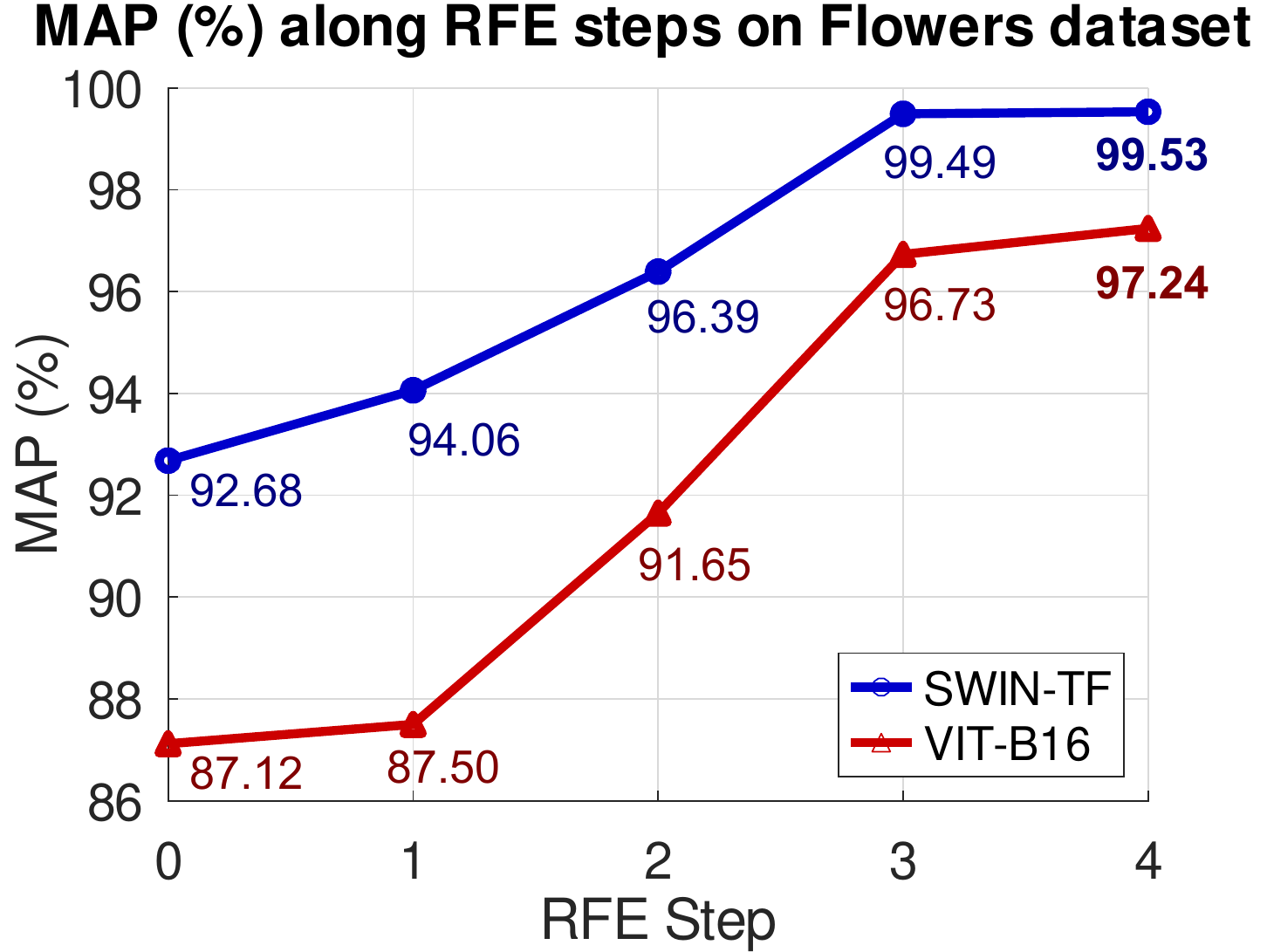}
\label{flowers_ablation}}
\hspace{0.20cm}
\subfloat[MPEG-7]
{\includegraphics[width=.34\textwidth]{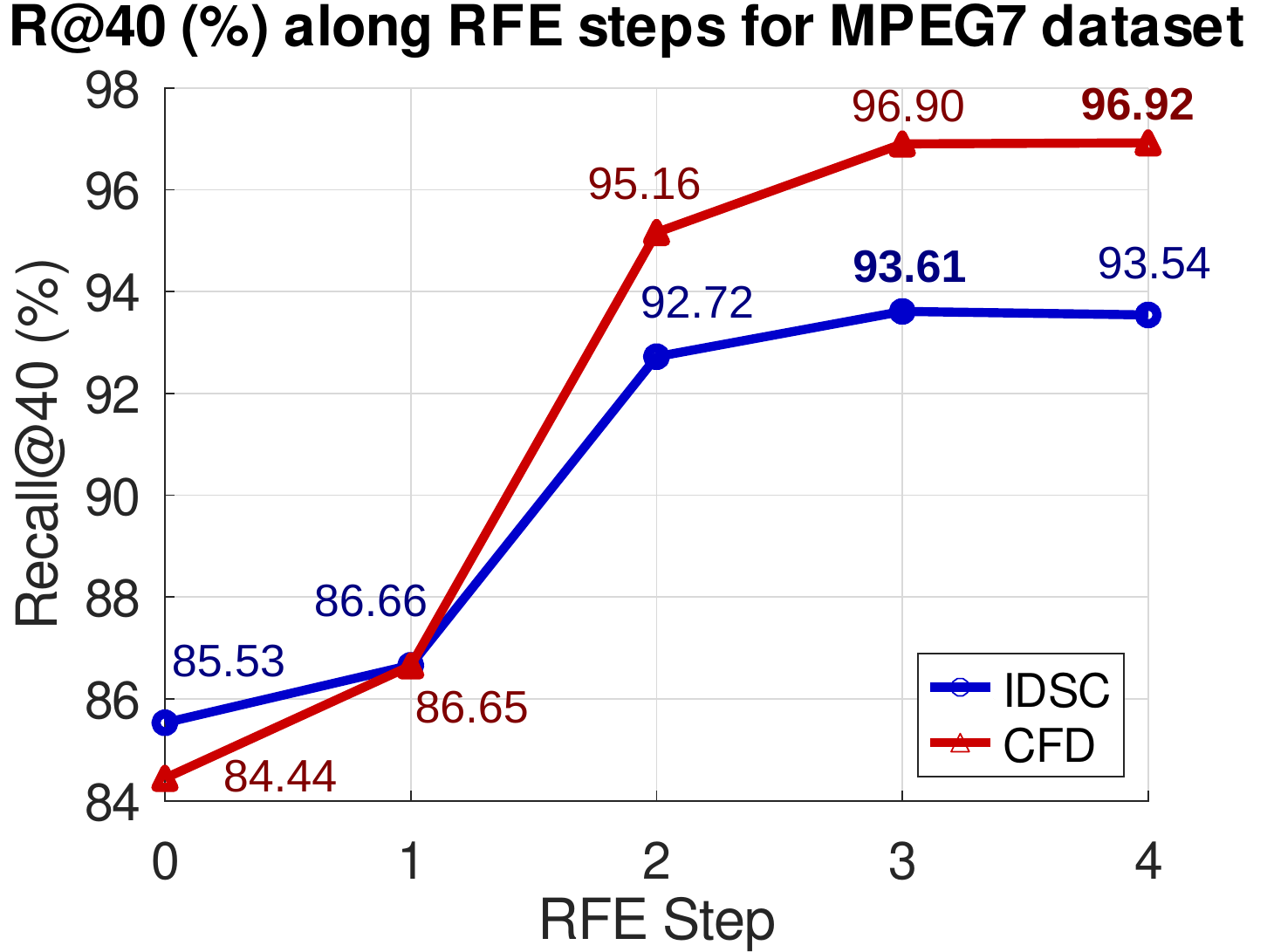}%
\label{mpeg7_ablation}}
\\
\vspace{0.15cm}
\subfloat[Holidays]
{\includegraphics[width=.34\textwidth]{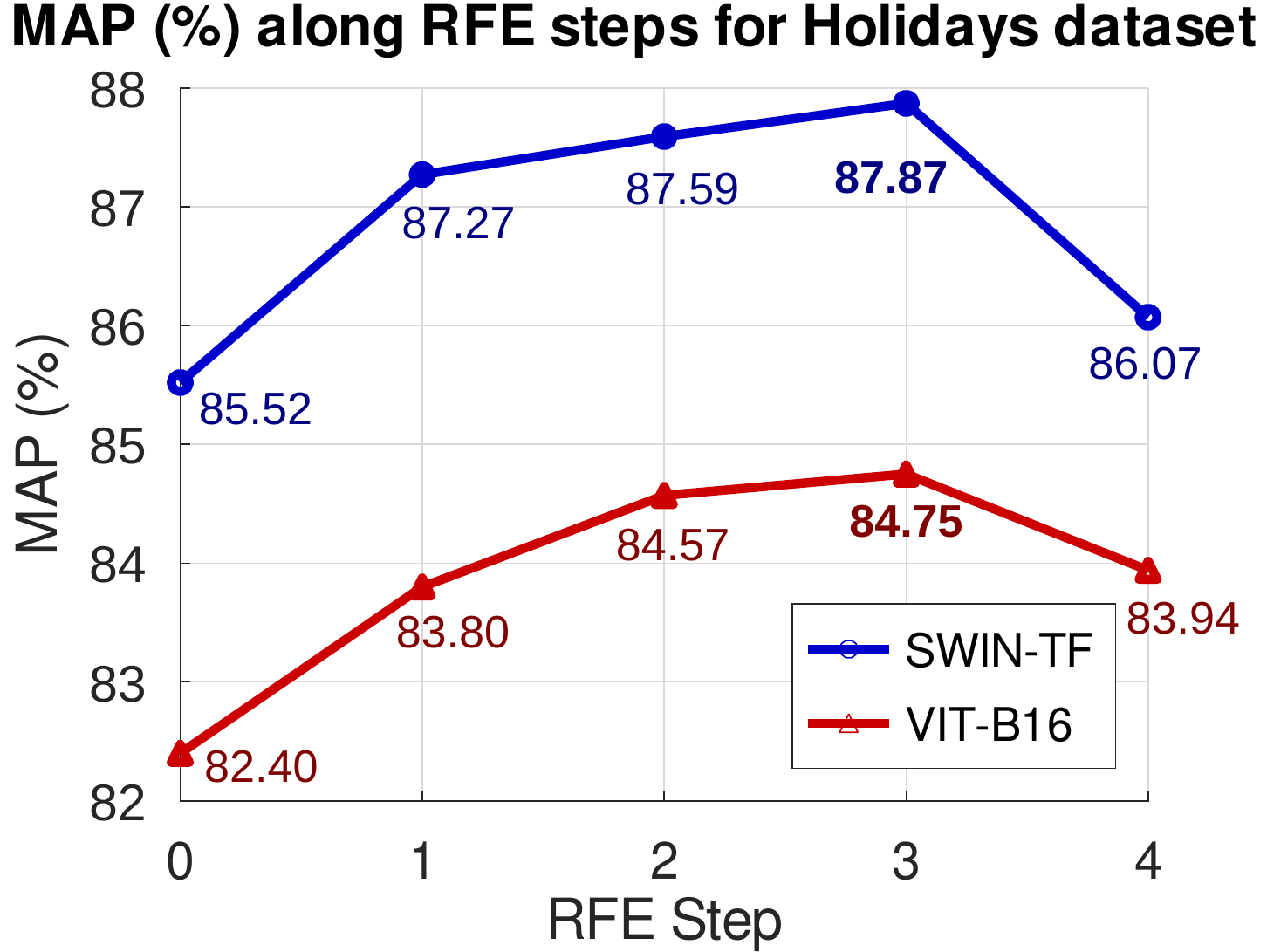}
\label{holidays_ablation}}
\hspace{0.20cm}
\subfloat[Corel5k]
{\includegraphics[width=.34\textwidth]{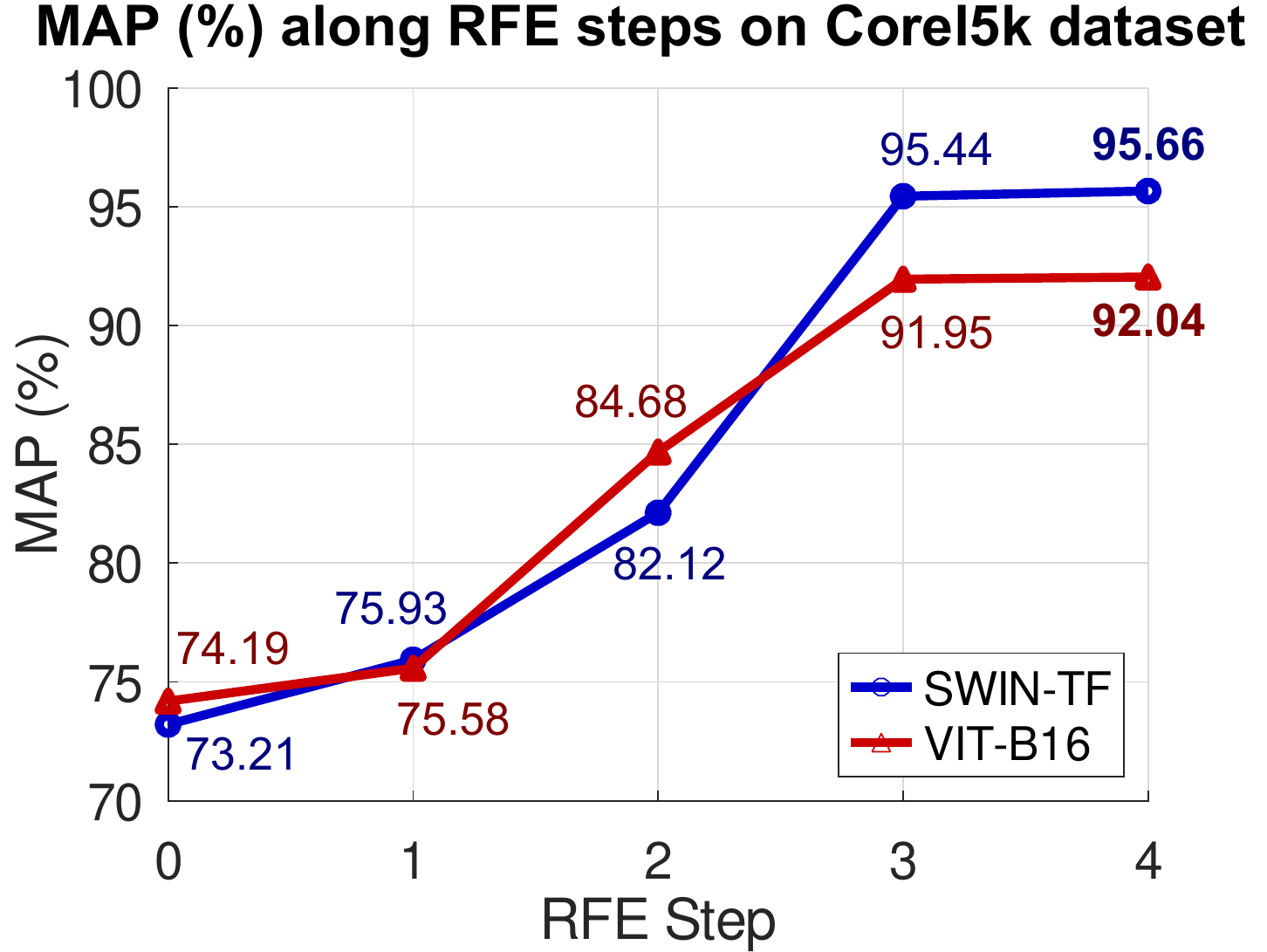}
\label{corel_ablation}}
\\
\vspace{0.15cm}
\subfloat[UKBench]
{\includegraphics[width=.34\textwidth]{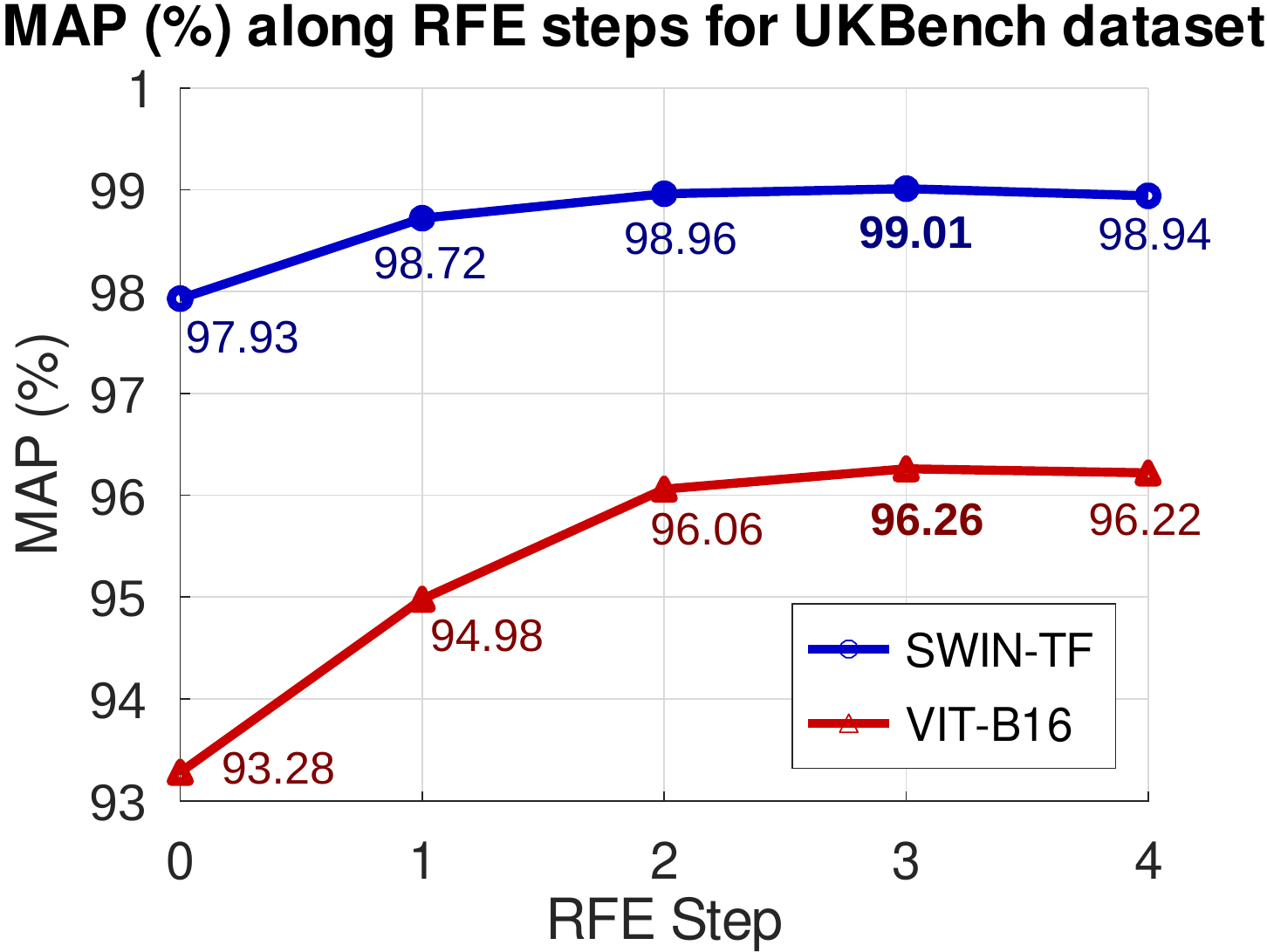}
\label{ukbench_ablation}}
\hspace{0.20cm}
\subfloat[CUHK03]
{\includegraphics[width=.34\textwidth]{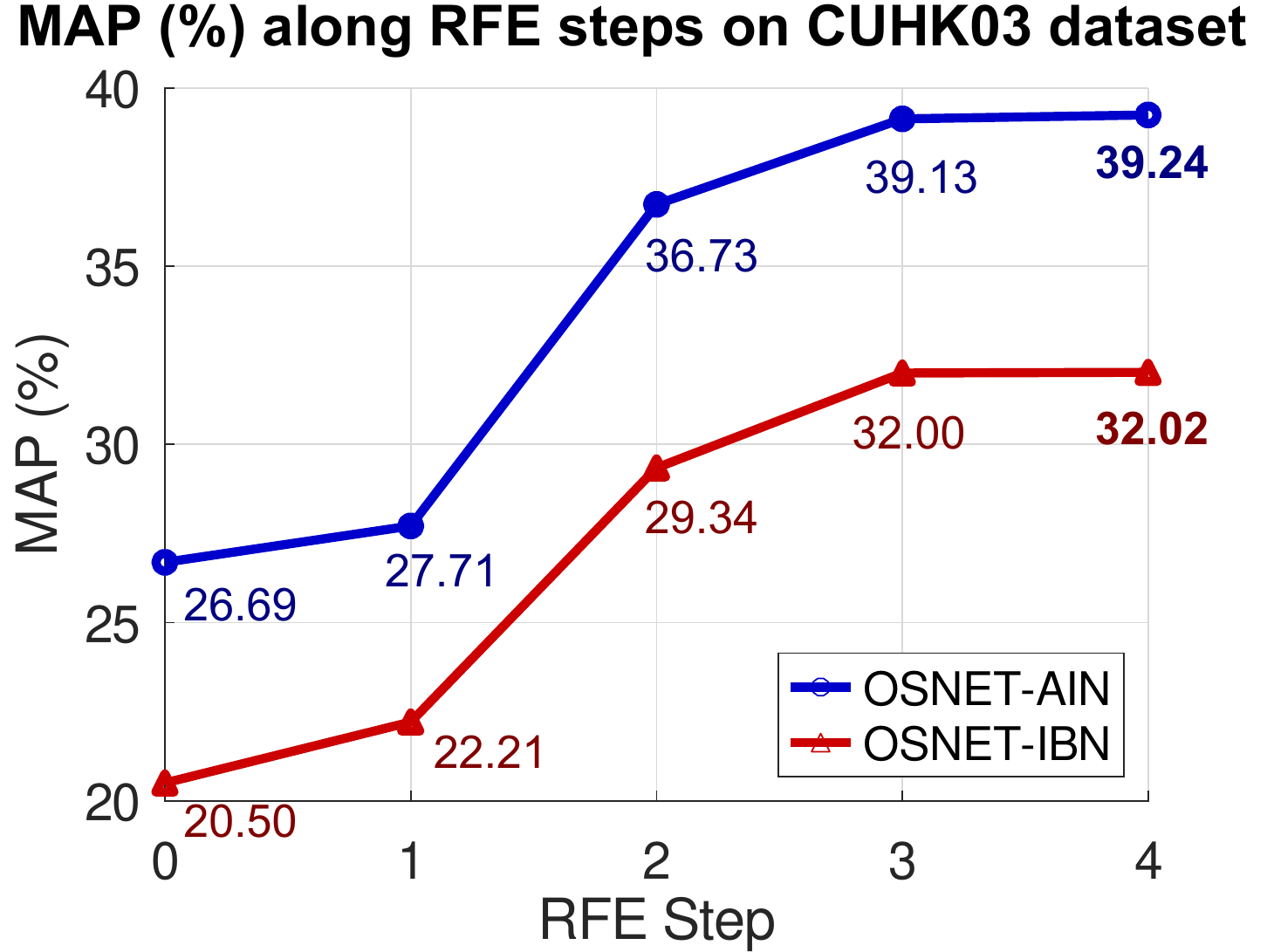}
\label{cuhk03_ablation}}
\caption{Ablation study on six datasets considering two descriptors each. The graphs present the effectiveness values (MAP or R@40 depending on the dataset) for each step of the proposed approach.
The best value for each plot is highlighted in bold.}
\label{fig:ablation_study}
\end{figure*}

Figure~\ref{fig:ablation_study} presents the effectiveness results for every step of the proposed approach.
For each dataset, two descriptors were evaluated. The descriptors considered were SWIN-TF~\cite{paperSWIN-TF}, VIT-B16~\cite{paperVIT16}, Inner Distance Shape
Context (IDSC)~\cite{PaperIDSC_Jacobs_2007}, Contour Features Descriptor (CFD)~\cite{PaperVISAPP2010}, OSNET-AIN~\cite{paperOSNET-IBN-AIN}, and OSNET-IBN~\cite{paperOSNET-IBN-AIN}; which are among the top-performing ones.
The experiment was conducted using the best value of $k$ in each case.
Notice that the values consistently increase along the performed steps, indicating the relevance of each step.
However, the datasets Holidays and Ukbench (c and e) revealed a different behavior, where Step 4 slightly decreases the MAP.
This is probably caused by the fact that different from others, these datasets have a small number of images per class.
Therefore, all the subsequent retrieval results presented in the next sections include Steps 1-4, except for UKBench and Holidays datasets, which use Steps 1-3.

%....................................................
\subsection{Retrieval Results}
\label{sec:RetResults}
%....................................................

In image retrieval tasks, there are two different scenarios, which are both included in our evaluation: \emph{(i)} standard re-ranking, where only one descriptor (feature) is considered; and \emph{(ii)} rank-aggregation, which combines one or more features.
For all experiments, we considered two variations for the parameter $k$ (size of the neighborhood set): a default value~\footnote{The default values are: $k=60$ for Flowers and Corel5k; $k=5$ for Holidays and UKBench; and $k=20$ for all the others.} and the best value.
The best $k$ is reported considering the executions with $k$ in range $[5, 120]$ with increments of 5.
In general, the results revealed that our method is very robust to the change of $k$.

Firstly, we evaluate RFE on Flowers, Corel5k, and ALOI datasets; which are general-purpose image datasets that use the same protocol and evaluation measure.
Table~\ref{tab:results_datasets1} presents the results.
For standard re-raking, a relative gain was reported considering the improvement in relation to the original input descriptor.
Since many descriptors are combined in rank aggregation, a gain is not reported in these scenarios.
Notice that for all the cases, significant gains were obtained (up to +50.84\%), and the fusion was able to improve the results even further.
The best result for each dataset is highlighted in bold and marked with a gray background.
For the three datasets, the best MAP is above 95\%.

\begin{table}[!ht]
\centering
\caption{Retrieval results of the proposed method (RFE) on general purpose \textbf{image datasets (Flowers, Corel5k, and ALOI)}. The results are reported for \textbf{MAP (\%)} evaluation measure considering re-ranking (single descriptor) and rank-aggregation (fusion of descriptors). The best values for each dataset are highlighted in bold with a gray background.}
\label{tab:results_datasets1}
\resizebox{.5\textwidth}{!}{
\begin{tabular}{lcccc}
\hline
 \textbf{Descriptors}   & \textbf{Original} & \textbf{Method w/} &  \textbf{Method w/}  & \textbf{Relative}   \\
    & \textbf{MAP} & \textbf{default $k$} & \textbf{best $k$}  & \textbf{Gain} \\ \hline
 \multicolumn{5}{c}{\textbf{Flowers}}                                                                   \\ \hline
 \multicolumn{5}{c}{\textbf{Re-Ranking}}                                                                   \\
 CNN-DPNet~\cite{paperCNN_DPN2017}    & 49.06     & 69.47  & 69.95 ($k$=70)  & +42.58\% \\
 CNN-ResNet~\cite{paperRESNET}    & 50.00      & 72.32  & 72.62 ($k$=75)  & +45.23\% \\
 CNN-SENet~\cite{paperCNN_SENET_2018}     & 40.85     & 61.26 & 61.26 ($k$=60)  & +49.96\% \\
 CNN-Xception~\cite{paperCNN_XCEPTION_2017} & 45.27     & 66.65  & 66.81 ($k$=65)  & +47.57\% \\
 T2T-VIT24T~\cite{paperT2T}    & 38.03     & 54.99  & 55.03 ($k$=70)  & +44.73\% \\
 VIT-B16 (VIT)~\cite{paperVIT16}      & 87.12     & 92.28  & 97.24 ($k$=80)  & +11.61\% \\
 SWIN-TF (STF)~\cite{paperSWIN-TF}      & 92.68     & 97.96  & 99.53 ($k$=85)  & +7.39\% \\
  \multicolumn{5}{c}{\textbf{Rank-Aggregation}}                                                                   \\
ResNet+DPNet            & --- & 80.07    & 80.13 ($k$=75) & ---     \\
 VIT+ResNet         & ---      & 94.63    & 97.67 ($k$=80) & ---     \\
 VIT+STF & \cellcolor{lightgray} --- &  \cellcolor{lightgray} \textbf{98.07}  & \cellcolor{lightgray} \textbf{99.65 ($k$=85)} & \cellcolor{lightgray} ---     \\
 VIT+ResNet+STF  & --- & 97.64  & 99.28 ($k$=90) & ---    \\
 \hline
 \multicolumn{5}{c}{\textbf{Corel5k}}                                                                   \\ \hline
\multicolumn{5}{c}{\textbf{Re-Ranking}}                                                                   \\
 CNN-DPNet~\cite{paperCNN_DPN2017}     & 63.69     & 81.58  & 85.48 ($k$=100) & +34.22\% \\
 CNN-ResNet~\cite{paperRESNET}    & 63.46     & 84.11 & 87.97 ($k$=100) & +38.61\% \\
 CNN-SENet~\cite{paperCNN_SENET_2018}     & 55.57     & 78.77  & 83.38 ($k$=100) & +50.06\% \\
 CNN-Xception~\cite{paperCNN_XCEPTION_2017}  & 52.92     & 76.33  & 79.82 ($k$=90)~  & +50.84\% \\
 T2T-VIT24T~\cite{paperT2T}    & 58.97     & 80.46  & 84.10 ($k$=100)  & +42.62\% \\
 VIT-B16 (VIT)~\cite{paperVIT16}       & 74.19     & 90.02 & 92.04 ($k$=100) & +24.06\% \\ 
 SWIN-TF (STF)~\cite{paperSWIN-TF}      & 73.21     & 93.55  & 95.66 ($k$=105)  & +30.70\% \\
\multicolumn{5}{c}{\textbf{Rank-Aggregation}}                                                                   \\
 ResNet+DPNet   & ---     & 87.66    & 91.22 ($k$=100) & ---    \\
 VIT+ResNet  & ---           & 93.28    & 95.01 ($k$=100) & ---    \\
 VIT+STF  & \cellcolor{lightgray} --- & \cellcolor{lightgray} \textbf{95.39}  & \cellcolor{lightgray} \textbf{96.79 ($k$=100)} & \cellcolor{lightgray} ---    \\
VIT+ResNet+STF  & --- &  95.20   & 96.79 ($k$=100) & ---    \\
\hline
 \multicolumn{5}{c}{\textbf{ALOI}}                                                                   \\ \hline
  \multicolumn{5}{c}{\textbf{Re-Ranking}}                                                                   \\
 CNN-DPNet~\cite{paperCNN_DPN2017}     & 79.09     & 94.45  & 96.32 ($k$=30)  & +21.79\% \\
 CNN-ResNet~\cite{paperRESNET}    & 81.97     & 94.79  & 96.37 ($k$=30)  & +17.57\% \\
 CNN-SENet~\cite{paperCNN_SENET_2018}     & 78.41     & 93.91  & 95.87 ($k$=30)  & +22.27\% \\
 CNN-Xception~\cite{paperCNN_XCEPTION_2017}  & 76.07     & 93.40  & 95.36 ($k$=30)  & +25.36\% \\
 T2T-VT24T~\cite{paperT2T}    & 76.90      & 93.46  & 95.36 ($k$=30)  & +24.00\% \\
 VIT-B16 (VIT)~\cite{paperVIT16}       & 79.40      & 93.55  & 95.40 ($k$=30)   & +20.16\% \\
 SWIN-TF (STF)~\cite{paperSWIN-TF}      & 89.97     & 96.68  & 97.81 ($k$=30)  & +8.71\% \\
\multicolumn{5}{c}{\textbf{Rank-Aggregation}}                                                                   \\
 ResNet+DPNet      & --- & 95.71    & 97.06 ($k$=30) & ---     \\
 VIT+ResNet            & --- & 95.70    & 97.13 ($k$=30) & ---     \\
 VIT+STF  & --- & 96.07  & 97.53 ($k$=30) & ---   \\
 VIT+ResNet+STF & \cellcolor{lightgray} --- &  \cellcolor{lightgray} \textbf{96.59}   & \cellcolor{lightgray} \textbf{97.73 ($k$=30)} & \cellcolor{lightgray} ---    \\
\hline
\end{tabular}
}
\end{table}

The same set of experiments was conducted for two datasets commonly used as image retrieval benchmarks: Holidays and UKbench.
Since they have a small number of images per class, the best $k$ is reported considering all the executions with $k$ in the range $[1, 20]$ with increments of 1.
Tables~\ref{tab:results_holidays} and~\ref{tab:results_ukbench} present the results for Holidays and Ukbench, respectively.
As can be seen, expressive gains were obtained for both datasets and measures.
For single descriptor executions, positive gains were obtained in all the cases, achieving gains up to +7.42\%.
For NS-Score, the results are very close to the maximum value, which is 4.
It is also possible to notice a correlation between MAP and NS-Score values.

\begin{table}[!ht]
\centering
\caption{Retrieval results of the proposed method (RFE) on the \textbf{Holidays dataset}. The results are reported for \textbf{MAP (\%)} evaluation measure considering re-ranking (single descriptor) and rank-aggregation (fusion of descriptors). The best values are highlighted in bold with a gray background.}
\label{tab:results_holidays}
\resizebox{.5\textwidth}{!}{
\begin{tabular}{lcccc}
\hline
 \textbf{Descriptors}   & \textbf{Original} & \textbf{Method w/} &  \textbf{Method w/}  & \textbf{Relative}   \\
    & \textbf{MAP} & \textbf{default $k$} & \textbf{best $k$}  & \textbf{Gain} \\ \hline
\multicolumn{5}{c}{\textbf{Re-Rank}}    \\
 CNN-DPNet~\cite{paperCNN_DPN2017}        & 70.58     & 74.64   & 75.00 ($k$=6)  & +6.25\% \\
 CNN-OLDFP~\cite{Mopuri_2015_CVPR_Workshops}       & 88.46     & 89.58   & 90.11 ($k$=6) & +1.87\% \\
 CNN-ResNet~\cite{paperRESNET}      & 74.87     & 77.15  & 77.37 ($k$=4) & +3.33\% \\
 CNN-SENet~\cite{paperCNN_SENET_2018}        & 71.59     & 74.36   & 74.36 ($k$=5) & +3.88\% \\
 CNN-Xception~\cite{paperCNN_XCEPTION_2017}    & 64.93     & 68.24   & 68.48 ($k$=6) & +5.46\% \\
 T2T-VIT24T~\cite{paperT2T}     & 69.04     & 73.98   & 74.03 ($k$=6) & +7.23\% \\
 VIT-B16 (VIT)~\cite{paperVIT16} & 82.40      & 84.75   & 84.75 ($k$=5) & +2.85\% \\
 SWIN-TF (STF)~\cite{paperSWIN-TF} & 85.52    & 87.87   & 87.87 ($k$=5) & +2.75\% \\
\multicolumn{5}{c}{\textbf{Rank-Aggregation}}    \\
 VIT+ResNet                 & --- & 86.11     & 86.22 ($k$=6) & ---     \\
 VIT+OLDFP                 & \cellcolor{lightgray} --- & \cellcolor{lightgray} \textbf{91.64}     & \cellcolor{lightgray} \textbf{91.97 ($k$=4)} & \cellcolor{lightgray} ---     \\
 ResNet+OLDFP                  & ---      & 88.08     & 88.33 ($k$=4) & ---     \\
 OLDFP+STF                & ---   & 90.84 & 90.88 ($k$=4) & --- \\
 VIT+ResNet+OLDFP      & --- & 89.98    & 90.35 ($k$=4)  & ---    \\
 VIT+OLDFP+STF     & ---  & 90.90    & 91.52 ($k$=4) & ---     \\
\hline
\end{tabular}
}
\end{table}

\begin{table*}[!ht]
\centering
\vspace{1cm}
\caption{Retrieval results of the proposed method (RFE) on the \textbf{UKBench dataset}. The results are reported for both \textbf{NS-Score and MAP} evaluation measures considering re-ranking (single descriptor) and rank-aggregation (fusion of descriptors). The best values are highlighted in bold with a gray background.}
\label{tab:results_ukbench}
\resizebox{\textwidth}{!}{
\begin{tabular}{l|cccc|cccc}
\hline
 \textbf{Evaluation Measure}   & \multicolumn{4}{c|}{\textbf{NS-Score}} &  \multicolumn{4}{c}{\textbf{MAP (\%)}}   \\ 
 \hline
 \textbf{Descriptors}   & \textbf{Original} & \textbf{Method w/} &  \textbf{Method w/}  & \textbf{Relative} & \textbf{Original} & \textbf{Method w/} &  \textbf{Method w/}  & \textbf{Relative}  \\
    & \textbf{NS-Score} & \textbf{default $k$} & \textbf{best $k$}  & \textbf{Gain} & \textbf{MAP} & \textbf{default $k$} & \textbf{best $k$}  & \textbf{Gain}\\ \hline
  & \multicolumn{4}{c|}{\textbf{Re-Ranking}} & \multicolumn{4}{c}{\textbf{Re-Ranking}}    \\
 CNN-DPNet~\cite{paperCNN_DPN2017}       & 3.46      & 3.71    & 3.72 ($k$=6) & +7.42\% & 90.47     & 94.58   & 94.67 ($k$=6) & +4.65\% \\
 CNN-OLDFP~\cite{Mopuri_2015_CVPR_Workshops}   & 3.85      & 3.93    & 3.93 ($k$=5) & +2.24\%     & 97.74     & 98.92  & 98.92 ($k$=5) & +1.21\% \\
 CNN-ResNet~\cite{paperRESNET}       & 3.67      & 3.85    & 3.85 ($k$=6) & +4.94\% & 94.54     & 97.31   & 97.31 ($k$=5) & +2.93\%  \\
 CNN-SENet~\cite{paperCNN_SENET_2018}       & 3.56      & 3.76   & 3.76 ($k$=5) & +5.52\% & 92.15     & 95.55  & 95.55 ($k$=5) & +3.69\% \\
 CNN-Xception~\cite{paperCNN_XCEPTION_2017}  & 3.49      & 3.75    & 3.75 ($k$=6) & +7.60\%  & 90.83     & 95.35   & 95.35 ($k$=6) & +4.99\% \\
 T2T-VIT24T~\cite{paperT2T}        & 3.48      & 3.75    & 3.75 ($k$=5) & +7.78\% & 90.26     & 95.40    & 95.40 ($k$=5)  & +5.69\% \\
 VIT-B16~\cite{paperVIT16}         & 3.62      & 3.80    & 3.80 ($k$=6)  & +5.00\% & 93.28     & 96.26   & 96.26 ($k$=5) & +3.19\% \\
 SWIN-TF~\cite{paperSWIN-TF}     & 3.86      & 3.94    & 3.94 ($k$=6)  & +2.01\% & 97.93      & 98.98   & 99.01 ($k$=6) & +1.10\% \\
  & \multicolumn{4}{c|}{\textbf{Rank-Aggregation}} & \multicolumn{4}{c}{\textbf{Rank-Aggregation}}    \\
VOC+OLDFP                       & --- & 3.90    & 3.90 ($k$=6) & --- & ---      & 98.22    & 98.22 ($k$=5) & ---     \\
 VOC+ResNet                     & --- & 3.92    & 3.93 ($k$=6)  & --- & ---         & 98.76     & 98.79 ($k$=6) & ---    \\
 VOC+VIT-B16                    & ---      & 3.92    & 3.92 ($k$=6) & --- & ---            & 98.69     & 98.77 ($k$=7) & ---     \\
 OLDFP+ResNet                   & --- & 3.94    & 3.95 ($k$=6)  & --- & ---   & 99.13    & 99.13 ($k$=5)  & ---     \\
 OLDFP+VIT-B16                  & ---     & 3.93    & 3.94 ($k$=6) & --- & ---      & 98.94    & 98.99 ($k$=6) & ---     \\
 ResNet+VIT-B16                 & ---    & 3.91    & 3.91 ($k$=5) & ---  & ---        & 98.45     & 98.45 ($k$=5) & ---    \\
 OLDFP+SWIN-TF                 & \cellcolor{lightgray} --- &  \cellcolor{lightgray} \textbf{3.97}     &  \cellcolor{lightgray} \textbf{3.97 ($k$=6)}  & \cellcolor{lightgray} --- & \cellcolor{lightgray} --- &  \cellcolor{lightgray} \textbf{99.53}      & \cellcolor{lightgray} \textbf{99.57 ($k$=6)} & \cellcolor{lightgray} ---    \\
 VOC+OLDFP+ResNet              & --- & 3.94  & 3.94 ($k$=6)   & --- & ---     & 99.07     & 99.07 ($k$=5) & ---     \\
 VOC+OLDFP+VIT-B16             & ---   & 3.94   & 3.95 ($k$=6) & --- & ---         & 99.09    & 99.13 ($k$=6) & ---    \\
 VOC+ResNet+VIT-B16            & ---        & 3.94    & 3.95 ($k$=6) & --- & ---         & 99.13    & 99.15 ($k$=6) & ---    \\
 OLDFP+ResNet+VIT-B16          & ---    & 3.94    & 3.94 ($k$=6) & --- & ---      & 99.07     & 99.08 ($k$=6) & ---      \\
 OLDFP+ResNet+SWIN-TF          & ---   & 3.96 & 3.96 ($k$=6) & --- & ---  & 99.40 & 99.41 ($k$=6) & --- \\
 VOC+OLDFP+ResNet+VIT-B16      & ---   & 3.95   & 3.95 ($k$=6) & --- & ---     & 99.20      & 99.28 ($k$=7)  & ---    \\
 VOC+OLDFP+VIT-B16+SWIN-TF     & ---  &  3.96     &  3.96 ($k$=6) & --- & ---  & 99.36      &  99.43 ($k$=6) & --- \\
\hline
\end{tabular}
}
\end{table*}

We also assessed RFE for person Re-ID (i.e., CUHK03, Market, and Duke datasets).
These datasets are usually more challenging.
They involve identifying and matching individuals across different camera views or even across different locations and times. People's appearances can vary significantly due to changes in lighting, pose, clothing, and accessories. These factors can make it difficult to match the same person in different images.
Table~\ref{tab:results_reid} reports the results on these datasets.
Since R1 is also commonly used for Re-ID evaluation, it was also included.
The R1 corresponds to the first value of the CMC (Cumulative
Matching Characteristics) curve, which indicates the number of ranked lists that
have an image that corresponds to the same individual in the first position after
the query image (which, in this case, is equivalent to Precision@1).
The best $k$ is reported considering all the executions with $k$ in the range $[5, 50]$ with increments of 5.
Notice that significant gains were obtained in all the cases (up to +65.88\%), which were also improved by the rank-aggregation in most cases.
These results reveal the potential of our approach in dealing not only with general-purpose scenarios but also with other challenging and more specific ones such as Re-ID.

\begin{table*}[!ht]
\centering
\caption{Retrieval results of the proposed method (RFE) on three person \textbf{Re-ID datasets (CUHK03, Market, and Duke)}. The results are reported for both \textbf{R1 and MAP} evaluation measures considering re-ranking (single descriptor) and rank-aggregation (fusion of descriptors).
The best values are highlighted in bold with a gray background (MAP as the criteria).
}
\label{tab:results_reid}
\resizebox{\textwidth}{!}{
\begin{tabular}{l|cccc|ccccc}
\hline
 \textbf{Evaluation Measure}   & \multicolumn{4}{c}{\textbf{R1 (\%)}} &  \multicolumn{4}{c}{\textbf{MAP (\%)}}   \\
    \hline
 \textbf{Descriptors}   & \textbf{Original} & \textbf{Method w/} &  \textbf{Method w/}  & \textbf{Relative} & \textbf{Original} & \textbf{Method w/} &  \textbf{Method w/}  & \textbf{Relative}   \\
    & \textbf{R1} & \textbf{default $k$} & \textbf{best $k$}  & \textbf{Gain} & \textbf{MAP} & \textbf{default $k$} & \textbf{best $k$}  & \textbf{Gain} \\ \hline
 &  \multicolumn{8}{c}{\multirow{2}{*}{\textbf{CUHK03}}}  \\ 
\\ \hline
  & \multicolumn{4}{c|}{\textbf{Re-Ranking}} & \multicolumn{4}{c}{\textbf{Re-Ranking}}    \\
 HACNN~\cite{paperHACNN}                 & 8.36      & 12.80  & 12.80 ($k$=20)  & +53.03\% & 9.33      & 14.27  & 14.41 ($k$=15) & +54.42\% \\
 MLFN~\cite{paperMLFN}                  & 9.47      & 13.69  & 13.79 ($k$=15) & +45.63\% & 9.85      & 15.14  & 15.18 ($k$=15) & +54.11\% \\
 OSNet-AIN~\cite{paperOSNET-IBN-AIN} & \cellcolor{lightgray} 26.39     & \cellcolor{lightgray} \textbf{36.67}  & \cellcolor{lightgray} \textbf{36.89 ($k$=15)} & \cellcolor{lightgray} +39.76\% & \cellcolor{lightgray} 26.69     & \cellcolor{lightgray} \textbf{39.12}  & \cellcolor{lightgray} \textbf{39.24 ($k$=15)} & \cellcolor{lightgray} +47.00\% \\
 OSNet-IBN~\cite{paperOSNET-IBN-AIN} & 20.31     & 29.65  & 29.82 ($k$=15) & +46.85\%        & 20.50      & 31.94  & 32.02 ($k$=15) & +56.18\% \\
 ResNet50~\cite{paperRESNET}    & 12.24     & 17.84  & 18.37 ($k$=15) & +50.15\%   & 12.74     & 19.77  & 19.77 ($k$=20) & +55.18\%  \\
  & \multicolumn{4}{c|}{\textbf{Rank-Aggregation}} & \multicolumn{4}{c}{\textbf{Rank-Aggregation}}    \\
 OSNet-AIN+OSNet-IBN              &  ---      &  36.19    &  37.16 ($k$=15) &  --- &  ---    &  38.51    & 39.13 ($k$=15) &  ---    \\
 OSNet-AIN+ResNet50               & ---   & 33.54     & 33.54 ($k$=20) & --- & ---               & 35.40     & 35.40 ($k$=20) & ---    \\
 OSNet-IBN+ResNet50               & ---           & 29.56    & 29.56 ($k$=20) & --- & ---                   & 31.40    & 31.40 ($k$=20) & ---     \\
 OSNet-AIN+OSNet-IBN+ResNet50     & --- & 33.91    & 33.91  ($k$=20)   & ---  & ---  & 35.94     & 35.94 ($k$=20) & ---    \\ \hline
 &  \multicolumn{8}{c}{\multirow{2}{*}{\textbf{Market}}}  \\ 
\\ \hline
  & \multicolumn{4}{c|}{\textbf{Re-Ranking}} & \multicolumn{4}{c}{\textbf{Re-Ranking}}    \\
 HACNN~\cite{paperHACNN}       & 49.23     & 52.20   & 52.82 ($k$=15) & +7.30\%          & 22.29     & 31.93  & 32.10 ($k$=25)  & +44.02\% \\
 MLFN~\cite{paperMLFN}          & 46.59     & 49.58  & 49.76 ($k$=15) & +6.82\%         & 21.11     & 30.65  & 30.89 ($k$=25) & +46.30\% \\
 OSNet-AIN~\cite{paperOSNET-IBN-AIN} & 69.95     & 70.99  & 70.99 ($k$=20) & +1.49\% & 42.33     & 57.38  & 58.21 ($k$=25) & +37.52\% \\
 OSNet-IBN~\cite{paperOSNET-IBN-AIN}  & 66.45     & 67.25  & 67.90 ($k$=15)  & +2.19\%      & 36.31     & 52.71   & 53.23 ($k$=25) & +46.60\% \\
 ResNet50~\cite{paperRESNET}         & 46.59     & 51.72 & 51.90 ($k$=15)  & +11.41\%      & 21.92     & 34.09  & 34.81 ($k$=25) & +58.82\% \\ 
  & \multicolumn{4}{c|}{\textbf{Rank-Aggregation}} & \multicolumn{4}{c}{\textbf{Rank-Aggregation}}    \\
 OSNet-AIN+OSNet-IBN           & \cellcolor{lightgray} ---   & \cellcolor{lightgray} \textbf{72.42}     & \cellcolor{lightgray} \textbf{72.42 ($k$=20)} & \cellcolor{lightgray} --- & \cellcolor{lightgray} ---                   & \cellcolor{lightgray} \textbf{58.55}    & \cellcolor{lightgray} \textbf{59.51 ($k$=25)} & \cellcolor{lightgray} ---   \\
 OSNet-AIN+ResNet50            & ---   & 67.34     & 67.34 ($k$=20) & --- & ---                         & 52.19    & 52.88 ($k$=25) & ---    \\
 OSNet-IBN+ResNet50            & ---   & 64.61     & 64.61 ($k$=20) & --- & ---                               & 49.45    & 50.40 ($k$=25) & ---       \\
 OSNet-AIN+OSNet-IBN+ResNet50  & ---   & 68.20      & 68.53 ($k$=15)  & --- & ---             & 54.35    & 55.11 ($k$=25) & ---    \\ 
 \hline
 &  \multicolumn{8}{c}{\multirow{2}{*}{\textbf{Duke}}}  \\ 
\\ \hline
  & \multicolumn{4}{c|}{\textbf{Re-Ranking}} & \multicolumn{4}{c}{\textbf{Re-Ranking}}    \\
 HACNN~\cite{paperHACNN}                & 42.19     & 50.31 & 50.99 ($k$=25) & +20.85\%  & 24.37     & 39.32  & 40.42 ($k$=25) & +65.88\% \\
 MLFN~\cite{paperMLFN}                 & 48.65     & 56.06  & 56.73 ($k$=25) & +16.61\%  & 28.00      & 44.00  & 45.39 ($k$=25) & +62.13\% \\
 OSNet-AIN~\cite{paperOSNET-IBN-AIN} & 71.14     & 75.67  & 76.84 ($k$=25) & +8.01\% & 51.68     & 66.60  & 68.31 ($k$=30) & +32.19\% \\
 OSNet-IBN~\cite{paperOSNET-IBN-AIN}        & 67.41     & 73.88  & 75.00 ($k$=25)  & +11.25\% & 44.66     & 63.60  & 64.81 ($k$=25) & +45.12\% \\
 ResNet50~\cite{paperRESNET}      & 52.29     & 60.50   & 62.57 ($k$=30) & +19.66\% & 31.00      & 48.77 & 50.67 ($k$=25) & +63.45\% \\
  & \multicolumn{4}{c|}{\textbf{Rank-Aggregation}} & \multicolumn{4}{c}{\textbf{Rank-Aggregation}}    \\
 OSNet-AIN+OSNet-IBN      & \cellcolor{lightgray} ---   & \cellcolor{lightgray} \textbf{76.21}    & \cellcolor{lightgray} \textbf{77.69 ($k$=25)}  & \cellcolor{lightgray} --- & \cellcolor{lightgray} ---                      & \cellcolor{lightgray} \textbf{67.46}    & \cellcolor{lightgray} \textbf{69.21 ($k$=25)}  & \cellcolor{lightgray} ---    \\
 OSNet-AIN+ResNet50       & ---   & 72.80      & 74.55 ($k$=30)   & --- & ---                      & 63.71    & 65.50 ($k$=25) & ---     \\
 OSNet-IBN+ResNet50       & ---   & 72.26     & 74.10 ($k$=30)   & --- & ---                         & 62.65    & 64.09 ($k$=25)  & ---    \\
 OSNet-AIN+OSNet-IBN+ResNet50  & ---  & 74.69   & 76.17 ($k$=25)  & --- & ---    & 65.74   & 67.02 ($k$=30) & ---   \\
 \hline
\end{tabular}
}
\end{table*}

%....................................................
\subsection{Classification Results}
\label{sec:clasifResults}
%....................................................

The proposed approach is capable of generating embeddings that can be utilized in various applications beyond retrieval.
In this section, we employ RFE for semi-supervised classification on two general-purpose image datasets (i.e., Flowers and Corel5k).
The process of embedding generation is unsupervised and encompasses all the steps of the proposed approach (from 1 to 5).
Our hypothesis is that the RFE embeddings can be used to train semi-supervised classifiers, resulting in improved accuracy.
We employed very recent Graph Convolutional Neural Networks (GCNs) models along with the traditional Support Vector Machine (SVM) with a polynomial kernel.
The GCNs can operate on graphs, and they have become increasingly popular due to their ability to handle complex relationships between data points, which cannot be easily modeled using traditional machine learning methods.

Tables~\ref{tab:semi_sup_flowers} and~\ref{tab:semi_sup_corel} present the results on Flowers and Corel5k datasets, respectively.
In all the classifiers, the default parameters were used, proposed by the original authors.
The GCNs were trained considering 50 epochs and $k=40$ for the input $k$NN graphs.
Our study compares the accuracy of classifiers that used the original features with those that used embeddings generated by the proposed RFE.
We highlight in bold the best result for each classifier and in red the best for each dataset.
The results demonstrate that the embeddings generated by our proposed approach are effective and have the potential to improve results across various classifiers. Notably, positive gains were obtained for all methods and features.

\begin{table*}[!th]
\centering
\caption{Semi-supervised classification (\textbf{accuracy}) on \textbf{Flowers} dataset for different features. We compare the training that used the original features with the one that used embeddings generated by the proposed RFE. The best result for each classifier is highlighted in bold and the best for each dataset is highlighted in red.}
\label{tab:semi_sup_flowers}
\begin{tabular}{clcccccc}
\hline
 \textbf{Mode}   & \textbf{Descriptor} & \textbf{SVM~\cite{paperSVM}} & \textbf{GCN-Net~\cite{paperGCN-ICLR2017}} & \textbf{GCN-Gat~\cite{paperGCN-GAT-ICLR2018}}  & \textbf{GCN-SGC~\cite{paperGCN-SGC2019}} & \textbf{GCN-APPNP~\cite{paperGCN-APPNP2019}} & \textbf{GCN-ARMA~\cite{paperGCN-ARMA2021}} \\ \hline
   & \textbf{CNN-ResNet~\cite{paperRESNET}}    & 82.467\% & 69.386\% & 71.211\%  & 78.649\% & 72.186\%      & 60.475\%  \\
      \textbf{Original} & \textbf{CNN-DPNet~\cite{paperCNN_DPN2017}}   & 79.812\% & 72.954\% & 18.874\%  & 76.292\% & 70.539\%      & 56.539\%  \\ 
     & \textbf{CNN-SENet~\cite{paperCNN_SENET_2018}} & 76.193\% & 68.895\% & 63.18\%   & 72.835\% & 66.797\%      & 60.649\%  \\ \hline
   \textbf{Our} & \textbf{CNN-ResNet~\cite{paperRESNET}}  & \textbf{82.565\%} & \textbf{82.593\%} & \textbf{82.966\%}  & \textcolor{red}{\textbf{84.948\%}} & \textbf{83.974\%}      & \textbf{75.160\%}   \\
   \textbf{Embeddings} &  \textbf{CNN-DPNet~\cite{paperCNN_DPN2017}} & 80.131\% & 80.003\% & 41.237\%  & 81.603\% & 81.029\%      & 67.784\%  \\
     & \textbf{CNN-SENet~\cite{paperCNN_SENET_2018}} & 76.716\% & 76.618\% & 73.454\%  & 77.559\% & 77.260\%       & 70.382\%  \\ \hline
   \textbf{Relative}  &  \textbf{CNN-ResNet~\cite{paperRESNET}}  & +0.12\%  & +19.03\% & +16.51\%  & +8.01\%  & +16.33\%      & +24.28\%  \\
     \textbf{Gain} &  \textbf{CNN-DPNet~\cite{paperCNN_DPN2017}} & +0.40\%   & +9.66\%  & +118.49\% & +6.96\%  & +14.87\%      & +19.89\%  \\
     & \textbf{CNN-SENet~\cite{paperCNN_SENET_2018}}   & +0.69\%  & +11.21\% & +16.26\%  & +6.49\%  & +15.66\%      & +16.05\%  \\
\hline
\end{tabular}
\end{table*}

\begin{table*}[!th]
\centering
\caption{Semi-supervised classification (\textbf{accuracy}) on \textbf{Corel5k} dataset for different features. We compare the training that used the original features with the one that used embeddings generated by the proposed RFE. The best result for each classifier is highlighted in bold and the best for each dataset is highlighted in red.}
\label{tab:semi_sup_corel}
\begin{tabular}{clcccccc}
\hline
 \textbf{Mode}   & \textbf{Descriptor} & \textbf{SVM~\cite{paperSVM}} & \textbf{GCN-Net~\cite{paperGCN-ICLR2017}} & \textbf{GCN-Gat~\cite{paperGCN-GAT-ICLR2018}}  & \textbf{GCN-SGC~\cite{paperGCN-SGC2019}} & \textbf{GCN-APPNP~\cite{paperGCN-APPNP2019}} & \textbf{GCN-ARMA~\cite{paperGCN-ARMA2021}} \\ \hline
    & \textbf{CNN-ResNet~\cite{paperRESNET}}  & 89.504\% & 78.066\% & 87.68\%   & 90.288\% & 86.679\%      & 73.621\%  \\
      \textbf{Original} &  \textbf{CNN-DPNet~\cite{paperCNN_DPN2017}} & 87.662\% & 84.733\% & 18.349\%  & 87.389\% & 85.653\%      & 72.883\%  \\
     & \textbf{CNN-SENet~\cite{paperCNN_SENET_2018}}   & 88.613\% & 88.627\% & 87.292\%  & 90.404\% & 88.76\%       & 83.447\%  \\ \hline
 \textbf{Our} & \textbf{CNN-ResNet~\cite{paperRESNET}} & \textbf{89.602\%} & 90.008\% & 91.003\%  & 91.54\%  & 91.507\%      & 89.212\%  \\
       \textbf{Embeddings} &  \textbf{CNN-DPNet~\cite{paperCNN_DPN2017}} & 87.933\% & 89.488\% & 52.374\%  & 90.515\% & 91.061\%      & 85.135\%  \\
& \textbf{CNN-SENet~\cite{paperCNN_SENET_2018}}   & 88.776\% & \textbf{91.299\%} & \textbf{91.441\%}  & \textbf{91.97\%}  & \textbf{\textcolor{red}{92.198\%}}      & \textbf{90.924\%}  \\ \hline
    \textbf{Relative} &  \textbf{CNN-ResNet~\cite{paperRESNET}}  & +0.11\%  & +15.3\%  & +3.79\%  &  +1.39\%  & +5.57\%   & +21.18\%  \\ 
      \textbf{Gain} &  \textbf{CNN-DPNet~\cite{paperCNN_DPN2017}} & +0.31\%  & +5.61\% & +185.43\% & +3.58\%  & +6.31\%       & +16.81\%  \\
     &  \textbf{CNN-SENet~\cite{paperCNN_SENET_2018}} & +0.18\%  & +3.01\%  & +4.75\%  & +1.73\%  & +3.87\%      & +8.96\%   \\
\hline
\end{tabular}
\end{table*}

%....................................................
\subsection{Unseen queries}
\label{sec:unseenQueries}
%....................................................

Encountering scenarios where query images are not included in the dataset being evaluated is not uncommon. These are referred to as external or unseen queries. To assess the proposed approach in such cases, we conducted experiments on the Flowers, Corel5k, and ALOI datasets, which are presented in Table~\ref{tab:uquery}.

We generated a set of unseen queries by randomly removing elements from the original dataset. To ensure a balanced analysis, we generated 10 samples per dataset, with each sample containing one element from each class. The reported MAP (both original and RFE) reflects the effectiveness of the approach in handling unseen queries, where the improvement is visible for all datasets and features.

\begin{table}[!ht]
\centering
\caption{Evaluation of RFE on \textbf{unseen queries} considering \textbf{MAP (\%)}. The reported results are the average of 10 executions, each conducted on a different set of unseen queries randomly sampled from the dataset.}
\label{tab:uquery}
\resizebox{.4\textwidth}{!}{
\begin{tabular}{llcc}
\hline
\textbf{Dataset} & \textbf{Descriptor} & \textbf{Original} & \textbf{RFE} \\ \hline
\multirow{3}{*}{\textbf{Flowers}}          & CNN-ResNet                           & 52.3226                              & 65.4526                          \\
                                  & VIT-B16                              & 89.0063                              & 93.3823                          \\
                                  & SWIN-TF                              & 93.0988                              & 95.3603                          \\ \hline

\multirow{3}{*}{\textbf{Corel5k}}          & CNN-ResNet                           & 63.2227                              & 76.3823                          \\
                                  & VIT-B16                              & 75.2124                              & 84.8642                          \\
                                  & SWIN-TF                              & 72.3914                             & 82.5962                           \\ \hline
\multirow{3}{*}{\textbf{ALOI}}             & CNN-ResNet                           & 82.5268                              & 88.4239                          \\
                                  & VIT-B16                              & 80.1258                              & 85.8109                          \\
                                  & SWIN-TF                              & 89.7562                              & 93.1862                          \\ \hline
\end{tabular}
}
\end{table}

%....................................................
\subsection{Comparison with State-of-the-art for Unsupervised Image Retrieval}
\label{sec:comparisonRetrieval}
%....................................................

This section aims to present the comparisons of the best results obtained by the proposed RFE (reported in Section~\ref{sec:RetResults}) in relation to the most recent baselines and state-of-the-art approaches on unsupervised image retrieval.

Table~\ref{tab:mpeg7} presents the results for ORL and MPEG-7 datasets, which are two traditional benchmark datasets.
These datasets are used for comparison with different diffusion methods.
The ORL consists of images of faces, while the MPEG-7 is composed of images of shapes and contours.
In order to keep consistency with the baselines, the same features were used for all the approaches: the IDSC~\cite{PaperIDSC_Jacobs_2007} for MPEG-7 and raw images for ORL. 
The result with the original features is reported as ``Our Baseline''.
The best values are highlighted in bold for each dataset.
Notice that RFE achieved the best result for ORL and comparable ones for MPEG-7.

\begin{table}[!ht]
\centering
\caption{\textbf{State-of-the-art (SOTA)} comparison with other variants of diffusion process on the \textbf{ORL (R@15)} and the \textbf{MPEG-7 (R@40)} datasets.}
\label{tab:mpeg7}
\resizebox{.35\textwidth}{!}{
\begin{tabular}{l|cc}
\hline
\textbf{Methods} & \textbf{ORL} & \textbf{MPEG-7} \\ \hline \hline
Baseline~\cite{paperRDP}  & 62.35 & 85.40  \\
SD~\cite{paperSD}        & 71.67 & 83.09  \\
LCDP~\cite{paperLCDP}      & 74.25 & 89.45  \\
TPG~\cite{paperTPG}       & 73.90 & 89.06  \\
MR~\cite{paperMR}        & 77.05 & 89.26  \\
MR*~\cite{paperMR}      & 77.58 & 92.61  \\
GDP~\cite{paperGDP}       & 77.42 & 90.96  \\
RDP (Y=I)~\cite{paperRDP} & 78.53 & \textbf{93.77}  \\
RDP (Y=W)~\cite{paperRDP} & 79.27 & \textbf{93.78}  \\ \hline
\multicolumn{1}{c|}{Our Baseline} & 74.32 & 85.40  \\
\multicolumn{1}{c|}{\cellcolor{lightgray}\textbf{RFE}}  & \cellcolor{lightgray}\textbf{90.62} & \cellcolor{lightgray}\textbf{93.54} \\ 
\multicolumn{1}{c|}{\cellcolor{lightgray}\textbf{(our method)}}  & \cellcolor{lightgray}\textbf{($k$=10)} & \cellcolor{lightgray}\textbf{($k$=20)} \\ \hline
\end{tabular}
}
\end{table}

The state-of-the-art comparison also encompasses the Flowers, Corel5k, and ALOI datasets; which is shown in Table~\ref{tab:baselines}.
Our method outperformed all other recent approaches, achieving the best results on all three datasets. The values reveal the effectiveness of RFE for both small and large datasets (Flowers and ALOI contain 13060 and 10200 images, respectively), with MAP always above 96.79\%.
This is a really significant result since the baselines also consider rank-aggregation of different features, especially Unsupervised Genetic Algorithm Framework for Rank Selection and fusion (UGAF-RSF)~\cite{paperUGAF-RSF} and Unsupervised Selective Rank Fusion (USRF)~\cite{paperUSRF} that combine more than 10 features.

\begin{table}[!ht]
\centering
\caption{\textbf{State-of-the-art (SOTA)} comparison on \textbf{Flowers, Corel5k, and ALOI} datasets \textbf{(MAP \%)}.}
\label{tab:baselines}
\resizebox{.41\textwidth}{!}{
\begin{tabular}{l|ccc}
\hline
\textbf{Method} & \textbf{Flowers} & \textbf{Corel5k} & \textbf{ALOI} \\ \hline \hline
CPRR~\cite{PaperCPRR} & --- & --- & 76.90 \\
RL-Sim~\cite{paperRLSim} & --- & --- & 78.84 \\
RL-Recom~\cite{paperICMR2015} & --- & --- & 80.35 \\

LHRR~\cite{paperLHRR} & --- & 73.34 & 88.42 \\ 
BFSTree~\cite{paperBFSTREE} & --- & 53.00 & 91.15 \\
RDPAC~\cite{paperRDPAC} & --- & 56.00 & 91.31 \\
UGAF-RSF~\cite{paperUGAF-RSF} & 80.92 & 91.45 & --- \\
USRF~\cite{paperUSRF} & 81.71 & 90.32 & --- \\ 
%\hline
\cellcolor{lightgray} \textbf{RFE (Our Method)} & \cellcolor{lightgray} \textbf{99.65} & \cellcolor{lightgray} \textbf{96.79} & \cellcolor{lightgray} \textbf{97.73} \\ \hline
\end{tabular}
}
\end{table}

%HOLIDAYS
\begin{table}[!ht]
\center
\caption{\textbf{State-of-the-art (SOTA)} comparison on \textbf{Holidays} dataset (\textbf{MAP}).}

 \label{tabDCompHolidays}
 
 \begin{tabular}{c|c|c|c|c}
 \hline
 \multicolumn{5}{c}{\textbf{MAP for state-of-the-art methods}  } \\
 	\hline
		J\'{e}gou  & Tolias & Paulin & Qin  & Zheng    \\
		\textit{et al.}~\cite{JegouECCV2008} & \textit{et al.}~\cite{ToliasICCV2013} & \textit{et al.}~\cite{PaperPathUnsup_IJVC2017} & \textit{et al.}~\cite{QinCVPR2013} & \textit{et al.}~\cite{ZhengTIP2014}   \\
	\hline			 		 		
		75.07\% &  82.20\% & 82.90\% & 84.40\% & 85.20\%    \\
		\hline
 \end{tabular}

\vspace{2mm}

\resizebox{8cm}{!} {
  \begin{tabular}{c|c|c|c|c}
  \hline
Sun & Zheng & Pedronette & Arandjelovic & Li \\
\textit{et al.}~\cite{PaperLocResSimReRank_INS2017} & \textit{et al.}~\cite{PaperMultiIndex_CVPR2014} & \textit{et al.}~\cite{PaperManLearReckNN_PR2017}  &  \textit{et al.}~\cite{paperNetVLAD} &  \textit{et al.}~\cite{LiCVPR2015} \\
 \hline
85.50\% & 85.80\% & 86.16\% & 87.50\% & 89.20\% \\
  \hline
  \end{tabular} 
}

\vspace{2mm}

 \resizebox{8cm}{!} {
  \begin{tabular}{c|c|c|c|c}
 \hline
Razavian & Pedronette & Gordo & Valem & Valem \\
\textit{et al.}~\cite{AliSRazavian2016} & \textit{et al.}~\cite{paperBFSTREE} & \textit{et al.}~\cite{paperGordoARL16a} & \textit{et al.}~\cite{paperUSRF}  & \textit{et al.}~\cite{paperRGSF} \\
 \hline
89.60\% & 90.02\% & 90.30\% & 90.51\% & 90.51\%  \\
  \hline
  \end{tabular} 
}

\vspace{2mm}

 \resizebox{8cm}{!} {
  \begin{tabular}{c|c|c|c|c}
 \hline
Liu & Pedronette & Pedronette & Yu & Berman \\
\textit{et al.}~\cite{LiCVPR2015} & \textit{et al.}~\cite{paperLHRR}  & \textit{et al.}~\cite{paperRDPAC} &  \textit{et al.}~\cite{paperYU2017235} & \textit{et al.}~\cite{paperMultiGrain2019}  \\
 \hline
90.89\% & 90.94\% & 91.25\% & 91.40\% & 91.80\% \\
  \hline
  \end{tabular} 
}

\vspace{2mm}

\resizebox{3cm}{!} {
\begin{tabular}{c}
\hline
\textbf{RFE (Our Method)}  \\
\hline
 \cellcolor{lightgray}\textbf{91.97\%} \\
\hline
\end{tabular} 
}
\end{table}

Tables~\ref{tabDCompHolidays} and~\ref{tabUKBenchStateArt} compare the RFE results to state-of-the-art methods on Holidays and Ukbench datasets, respectively.
These datasets are widely used as benchmarks for many retrieval algorithms.
We compare RFE to at least 15 approaches for each dataset.
Notice, that the results achieved by RFE are higher than the baselines in both cases.
We achieved a N-S Score of 3.97 (the maximum possible value is 4.00).

%UKBENCH
\begin{table}[ht!]
 \centering		
\caption{\textbf{State-of-the-art (SOTA)} comparison on \textbf{UKBench} dataset (\textbf{NS-Score}).}
 
 \label{tabUKBenchStateArt}

\resizebox{8cm}{!} {
 \begin{tabular}{c|c|c|c|c}
 	\hline
 	\multicolumn{5}{c}{\textbf{N-S-Scores for state-of-the-art methods}  } \\
 	\hline
 	Qin 	 &  Zhang   & Zheng  & Bai & Xie\\
 		
 		   \textit{et al.}~\cite{PaperHelloNeig_CVPR2011}   & \textit{et al.}~\cite{PaperQFusion_PAMI2015} & \textit{et al.}~\cite{zheng2015query} &  \textit{et al.}~\cite{PaperBai_TIP2016} & \textit{et al.}~\cite{PaperONEICMR2015} \\
 	\hline
 	 3.67   & 3.83 & 3.84 & 3.86 & 3.89 \\
   	\hline
 \end{tabular}
% }
 }

\vspace{2mm}

\resizebox{8cm}{!} {
 \begin{tabular}{c|c|c|c|c}	
\hline
    Lv  & Liu  & Pedronette  & Bai  & Liu  \\
	\textit{et al.}~\cite{paperYueNS2018}  & 	\textit{et al.}~\cite{PaperImgGraphRankLevFus_TIP2017}  & 	\textit{et al.}~\cite{PaperManLearReckNN_PR2017}  & 	\textit{et al.}~\cite{PaperEnsDiffICCV20017}  & 	\textit{et al.}~\cite{LAO2021103282}  \\
	 \hline
    3.91  & 3.92  & 3.93  & 3.93  & 3.93  \\
   	\hline
 \end{tabular}
 }

\vspace{2mm}

\resizebox{8cm}{!} {
 \begin{tabular}{c|c|c|c|c}
 \hline
    Valem  & Bai  & Valem  & Valem  & Chen  \\
	\textit{et al.}~\cite{paperRGSF}  & 	\textit{et al.}~\cite{BaiBTL17}  & 	\textit{et al.}~\cite{paperUSRF}  & 	\textit{et al.}~\cite{paperUGAF-RSF}  & 	\textit{et al.}~\cite{chen_ukbench_2020}  \\
	 \hline
    3.93  & 3.94  & 3.94  & 3.95  & 3.96  \\ 
   	\hline
 \end{tabular}
 }

\vspace{2mm}

\resizebox{3cm}{!} {
\begin{tabular}{c}
\hline
\textbf{RFE (Our Method)}  \\
\hline
\cellcolor{lightgray}\textbf{3.97} \\
\hline
\end{tabular} 
}
\end{table}

Table~\ref{tab:state_of_art_reid} presents the results of different approaches on the Re-ID datasets considering both R1 and MAP.
Our results (RFE) are marked with a gray background and correspond to the best ones according to Table~\ref{tab:results_reid}.
The abbreviations in parentheses indicate the datasets used for training (C03 = CUHK03, M = Market1501, D = DukeMTMC, MT = MSMT17).
For example, the use of (D, M) indicates that the reported result corresponds to training done either on Duke or on Market dataset. The results reported on Market were trained on Duke and the results reported on Duke were trained on Market.
None of the presented methods were trained using labels of the target dataset.
The abbreviations were omitted for multi-source baselines, but they can be consulted in their papers.
The best results for each dataset are highlighted in bold.
Notice, that our results are among the best in all the cases and are above all of the baselines for DukeMTMC considering MAP.
\begin{table}[!th]
\centering
\caption{\textbf{State-of-the-art (SOTA)} comparison for \textbf{person Re-ID} datasets considering \textbf{MAP} (\%) and \textbf{R-01} (\%).
The abbreviations in parentheses indicate the datasets used for training (C03 = CUHK03, M = Market1501, D = DukeMTMC, MT = MSMT17).
For example, the use of (D, M) indicates that the reported result corresponds to training done either on Duke or on Market dataset. The results reported on Market were trained on Duke and the results reported on Duke were trained on Market. None of the presented methods were trained using labels of the target dataset.
}
\label{tab:state_of_art_reid}
\resizebox{.49\textwidth}{!}{ 
\begin{tabular}{lc|cc|cc|cc}
\hline
%\cline{3-8}
                                    \multicolumn{1}{c}{} & \multicolumn{1}{c|}{}   & \multicolumn{6}{c}{\textbf{Datasets}}                                                                                                                                                                             \\ \cline{3-8} 
                              \multicolumn{1}{l}{\textbf{Method}} &   \multicolumn{1}{c|}{\textbf{Year}}      & \multicolumn{2}{c|}{\textbf{Market1501}}                             & \multicolumn{2}{c|}{\textbf{DukeMTMC}}                               & \multicolumn{2}{c}{\textbf{CUHK03}}                                 \\ \cline{3-8}
     \multicolumn{1}{c}{} & \multicolumn{1}{c|}{} & \multicolumn{1}{c}{\textbf{R1}} & \multicolumn{1}{c|}{\textbf{MAP}} & \multicolumn{1}{c}{\textbf{R1}} & \multicolumn{1}{c|}{\textbf{MAP}} & \multicolumn{1}{c}{\textbf{R1}} & \multicolumn{1}{c}{\textbf{MAP}} \\ \cline{3-8}
\hline
\multicolumn{8}{c}{\textbf{Unsupervised Methods}} \\ \hline
\hline
\multicolumn{1}{l}{ARN~\cite{paperARN}}    &  2018      & 70.3                             & 39.4                              & 60.2                             & 33.4                              & ---                              & ---                               \\ 
\multicolumn{1}{l}{EANet~\cite{paperEANET}}     & 2018    & 66.4                             & 40.6                              & 45.0                             & 26.4                              & 51.4                            & 31.7                              \\ 
\multicolumn{1}{l}{ECN~\cite{paperECN}}       &  2019   & 75.1                             & 43.0                              & 63.3                             & 40.4                              & ---                              & ---                               \\
\multicolumn{1}{l}{TAUDL~\cite{paperTAUDL}}   &  2018     & 63.7                             & 41.2                              & 61.7                             & 43.5                              & 44.7                            & 31.2                              \\
\multicolumn{1}{l}{UTAL~\cite{paperUTAL}}    & 2019  & 69.2                             & 46.2                              & 62.3                             & 44.6                              & \textbf{56.3}                            & \textbf{42.3}                              \\
\multicolumn{1}{l}{SSL~\cite{paperSSL}}    & 2020  & 71.7                             & 37.8                              & 52.5                             & 28.6 & ---                             & ---                              \\
\multicolumn{1}{l}{HCT~\cite{paperHCT}}    & 2020  & 80.0                             & 56.4                             & 69.6                             & 50.7 & ---                             & ---                              \\
\multicolumn{1}{l}{CAP~\cite{paperCAP}}    & 2021  & \textbf{91.4}                             & \textbf{79.2}                             & \textbf{81.1}                             & 67.3  & ---                             & ---                              \\
\multicolumn{1}{l}{IICS~\cite{paperIICS2021}}    & 2021  & 89.5 & 72.9 & 80.0 & 64.4  & ---                             & ---                              \\
\hline
\multicolumn{8}{c}{\textbf{Domain Adaptive Methods}} \\ \hline
\hline
\multicolumn{1}{l}{HHL (D,M)~\cite{paperHHL}}  & 2018   & 62.2                             & 31.4                              & 46.9                             & 27.2                              & ---                              & ---                               \\
\multicolumn{1}{l}{HHL (C03)~\cite{paperHHL}}  & 2018   & 56.8                             & 29.8                              & 42.7                             & 23.4                              & ---                              & ---                               \\ 
\multicolumn{1}{l}{ATNet (D,M)~\cite{paperATNET}} & 2019  & 55.7                             & 25.6                              & 45.1                             & 24.9                              & ---                              & ---                               \\
\multicolumn{1}{l}{CSGLP (D,M)~\cite{paperCSGLP}} & 2019  & 63.7                             & 33.9                              & 56.1                             & 36.0                              & ---                              & ---                               \\ 
\multicolumn{1}{l}{ISSDA (D,M)~\cite{paperISSDA}} & 2019  & 81.3                             & 63.1                             & 72.8                             & 54.1                              & ---                              & ---                               \\
\multicolumn{1}{l}{ECN++ (D,M)~\cite{paperECN++2020}}  & 2020 & 84.1 & 63.8 & 74.0 & 54.4 & ---                              & ---                               \\
\multicolumn{1}{l}{MMCL (D,M)~\cite{paperMMCL2020}}  & 2020 & 84.4 & 60.4 & 72.4 & 51.4 & ---                              & ---                               \\
\hline
\multicolumn{8}{c}{\textbf{Cross-Domain Methods (single-source)}} \\ \hline
\hline
\multicolumn{1}{l}{EANet (C03)~\cite{paperEANET}} & 2018  & 59.4                             & 33.3                              & 39.3                             & 22.0                              & ---                              & ---                               \\ 
\multicolumn{1}{l}{EANet (D,M)~\cite{paperEANET}} & 2018  & 61.7                             & 32.9                              & 51.4                             & 31.7                              & ---                              & ---                               \\ 
\multicolumn{1}{l}{SPGAN (D,M)~\cite{paperSPGAN}} &  2018  & 43.1                             & 17.0                              & 33.1                             & 16.7                              & ---                              & ---                               \\ 
\multicolumn{1}{l}{DAAM (D,M)~\cite{paperDAAM}} &  2019  & 42.3                             & 17.5                              & 29.3                             & 14.5                              & ---                              & ---                               \\ 
\multicolumn{1}{l}{AF3 (D,M)~\cite{paperAF3}} &  2019   & 67.2                             & 36.3                              & 56.8                             & 37.4                              & ---                              & ---                               \\ 
\multicolumn{1}{l}{AF3 (MT)~\cite{paperAF3}}  & 2019    & 68.0                             & 37.7                              & 66.3                             & 46.2                              & ---                              & ---                               \\ 
\multicolumn{1}{l}{PAUL (MT)~\cite{paperPAUL}}     & 2019 & 68.5                             & 40.1                              & 72.0                             & 53.2                              & ---                              & ---                               \\ 
\hline
\multicolumn{8}{c}{\textbf{Cross-Domain Methods (multi-source)}} \\ 
\hline
\multicolumn{1}{l}{EMTL~\cite{paperEMTL}} & 2018 & 52.8                             & 25.1                              & 39.7                             & 22.3                              & ---                              & ---                               \\ 
\multicolumn{1}{l}{CAMEL~\cite{paperCAMEL}}          & 2017 & 54.5                             & 26.3                              & ---                              & ---                               & 31.9                             & ---                               \\ 
\multicolumn{1}{l}{Baseline by~\cite{paperLargestMS}}    & 2019 & 80.5                             & 56.8                              & 67.4                             & 46.9                              & 29.4                             & 27.4                              \\ 
\hline
\multicolumn{8}{c}{\textbf{Our Proposed Method}} \\ \hline
\hline
\multicolumn{1}{l}{\textbf{Our Method}} &  & \cellcolor{lightgray}\textbf{72.42} & \cellcolor{lightgray}\textbf{59.51}  & \cellcolor{lightgray}\textbf{77.69} & \cellcolor{lightgray}\textbf{69.21}  & \cellcolor{lightgray}\textbf{36.89} & \cellcolor{lightgray}\textbf{39.24}  \\ \hline
\end{tabular}
}
\end{table}

%....................................................
\subsection{Comparison with State-of-the-art for Semi-Supervised Image Classification}
\label{sec:comparisonClassification}
%....................................................

This section compares the semi-supervised image classification results reported in Section~\ref{sec:clasifResults} to various state-of-the-art approaches.
Table~\ref{tab:baselines_classification} presents the comparisons considering different features (CNN-ResNet~\cite{paperRESNET} and CNN-SENet~\cite{paperCNN_SENET_2018}).
The best result for each feature and dataset is highlighted in bold.
The gray rows indicate the results that correspond to our method.
We employed the same protocol adopted for RFE in all baselines: 5 executions of 10 folds.
The only exception is CoMatch~\cite{paperCoMatch}, where only 3 executions were reported for Corel5k due to the long time required to train this approach.
Different from others, CoMatch takes images as input.
However, it uses CNN-ResNet as its backbone.

\begin{table}[!ht]
\centering
\caption{\textbf{Accuracy} comparison (\%) for baselines on \textbf{Flowers} and \textbf{Corel5k} datasets. We compared our approach with \textbf{semi-supervised classification baselines}. The methods are compared with different input features. The results of our method are highlighted with a gray background; the best results for each pair of features and dataset are marked in bold.}
\label{tab:baselines_classification}
\resizebox{.48\textwidth}{!}{ 
\begin{tabular}{l|c|c|c}
\hline
\textbf{Method} & \textbf{Input} & \textbf{Flowers} & \textbf{Corel5k} \\ \hline 
\hline
\textbf{CoMatch}~\cite{paperCoMatch} & ~\textbf{Images} &  82.55 & \emph{85.70} \\
\hline
\textbf{kNN} & & 63.67 & 76.80 \\
\textbf{SVM}~\cite{paperSVM} &  & 80.54 & 88.73 \\
\textbf{OPF}~\cite{paperSSOPF2014} & & 71.77 & 83.56\\
\textbf{SL-Perceptron}  & & 75.44 & 83.56 \\
\textbf{ML-Perceptron} & & 78.88 & 87.10 \\
\textbf{PseudoLabel+SGD}~\cite{lee2013pseudo} & & 82.69 & 89.76  \\
\textbf{LS+kNN}~\cite{Zhou04learningwith} & \textbf{ResNet} & 73.49 & 83.98  \\
\textbf{LS+SVM}~\cite{Zhou04learningwith, paperSVM} & \textbf{Features} & 73.53 & 83.26   \\
\textbf{LS+OPF}~\cite{Zhou04learningwith, paperSSOPF2014} & & 72.66 & 82.32 \\
\textbf{LS+SL-Perceptron}~\cite{Zhou04learningwith} & & 72.34 & 82.38   \\
\textbf{LS+ML-Perceptron}~\cite{Zhou04learningwith} & & 73.03 & 82.53  \\
\textbf{GNN-LDS}~\cite{paperLDS_GNN_2019} & & 54.98 & 62.69  \\
\textbf{GNN-KNN-LDS}~\cite{paperLDS_GNN_2019} &  & 79.32 & 88.94  \\
\textbf{WSEF}~\cite{paperWSJoao2020} & & \textbf{85.12}  & \textbf{91.68}  \\
\cellcolor{lightgray} \textbf{RFE (Our Method)} & & \cellcolor{lightgray} 84.95  & \cellcolor{lightgray} 91.54 \\
\hline

\textbf{kNN} & & 48.71 & 58.78 \\
\textbf{SVM}~\cite{paperSVM} & & 73.30 & 85.89 \\
\textbf{OPF}~\cite{paperSSOPF2014} & & 64.00 & 81.33 \\
\textbf{SL-Perceptron} & & 71.84 & 82.28  \\
\textbf{ML-Perceptron} & & 72.62 & 86.90  \\
\textbf{PseudoLabel+SGD}~\cite{lee2013pseudo} & & 76.87 & 89.85  \\
\textbf{LS+kNN}~\cite{Zhou04learningwith} & \textbf{SENet} & 58.05 & 72.16   \\
\textbf{LS+SVM}~\cite{Zhou04learningwith, paperSVM} & \textbf{Features} & 59.84  & 72.79  \\
\textbf{LS+OPF}~\cite{Zhou04learningwith, paperSSOPF2014} & & 59.25 & 72.20  \\
\textbf{LS+SL-Perceptron}~\cite{Zhou04learningwith} &  & 59.27 & 72.19   \\
\textbf{LS+ML-Perceptron}~\cite{Zhou04learningwith} & & 59.39  & 72.24  \\
\textbf{GNN-LDS}~\cite{paperLDS_GNN_2019} & & 52.24 &  65.80  \\
\textbf{GNN-KNN-LDS}~\cite{paperLDS_GNN_2019} & & 73.69 & 89.95  \\
\textbf{WSEF}~\cite{paperWSJoao2020} &  & 76.16  & 89.74  \\
\cellcolor{lightgray}\textbf{RFE (Our Method)} & & \cellcolor{lightgray} \textbf{77.56}  & \cellcolor{lightgray} \textbf{92.20} \\
\hline
\end{tabular}
}
\end{table}

For all the methods, we considered the default parameters and implementation provided by the original authors or the one in \emph{Python Sklearn}.
Regarding parameters, we used $k = 20$ for methods that require a size for the neighborhood set (i.e, kNN, GNN-LDS, GNN-KNN-LDS, and WSEF).
The Label Spreading (LS)~\cite{Zhou04learningwith} was used combined with different classifiers once it can be used to generate pseudo-labels for further expanding the training set.
The results achieved by RFE are the best ones for the SENet features and very comparable to the best for the ResNet features.

%..........................................
\subsection{Visual Analysis}
\label{sec:visual}
%..........................................

\begin{figure*}[!thb]
    \centering
    \begin{tabular}{ccc} 
    \includegraphics[width=.31\textwidth]{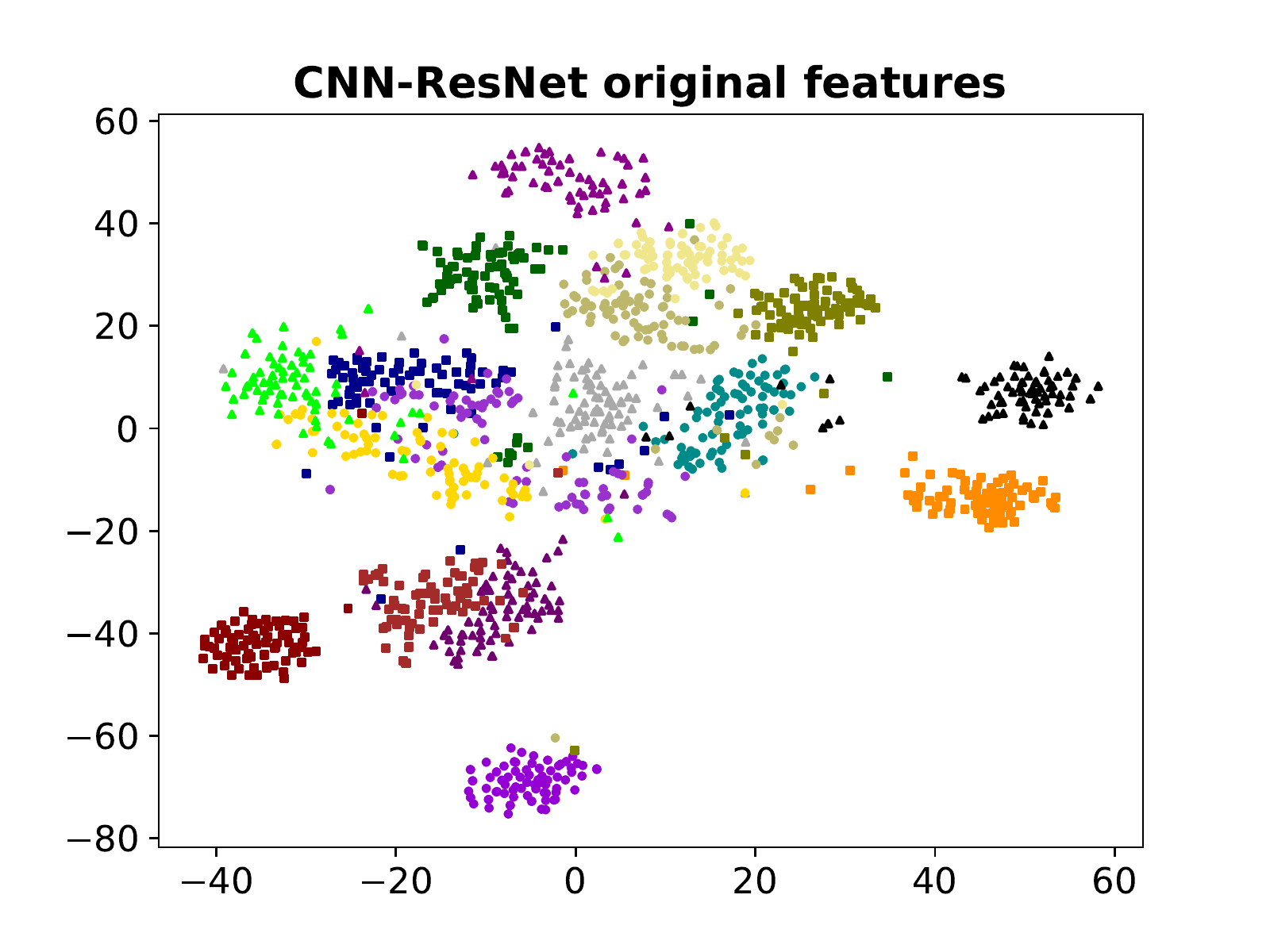} &
    \includegraphics[width=.31\textwidth]{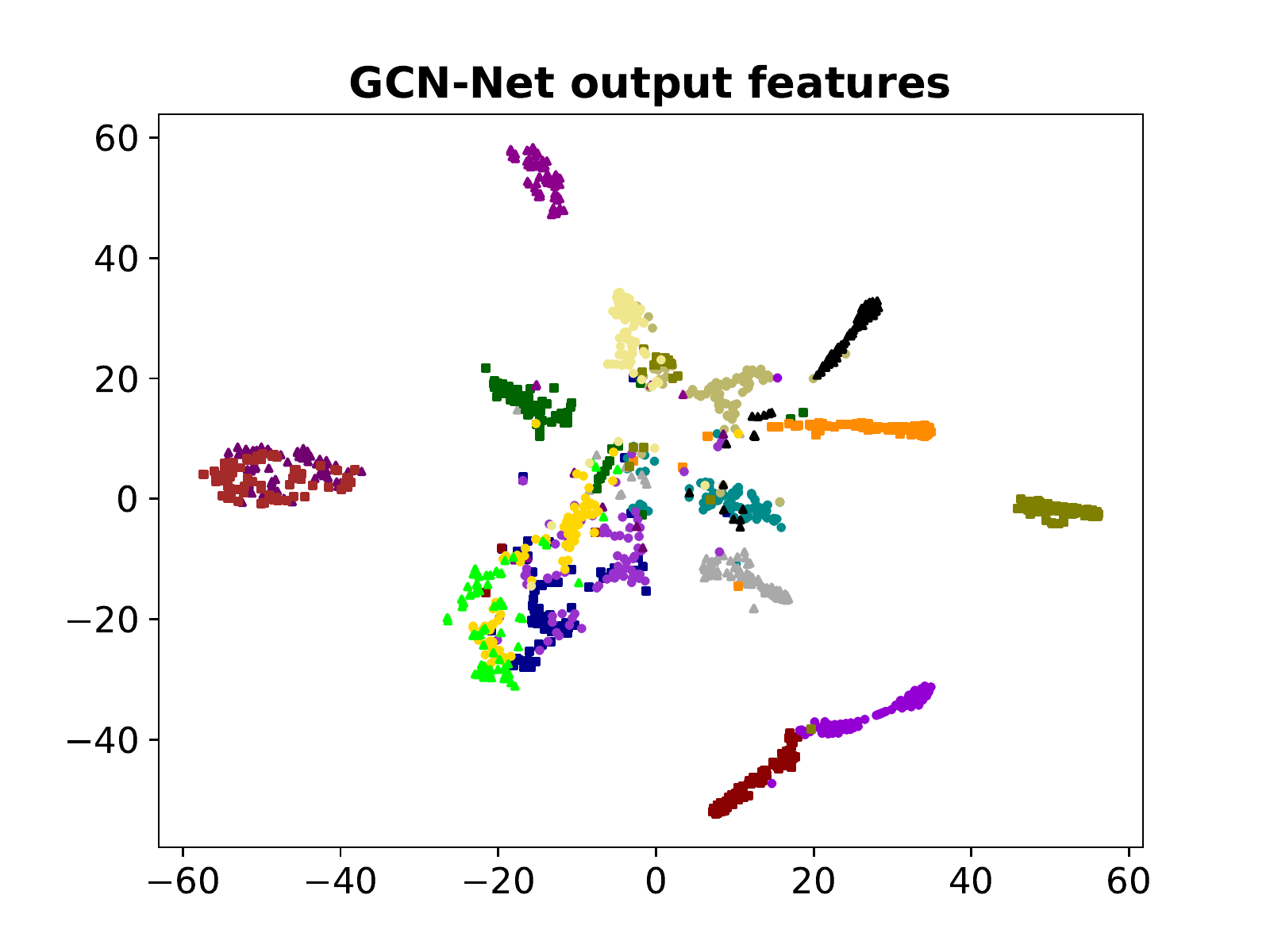} &
    \includegraphics[width=.31\textwidth]{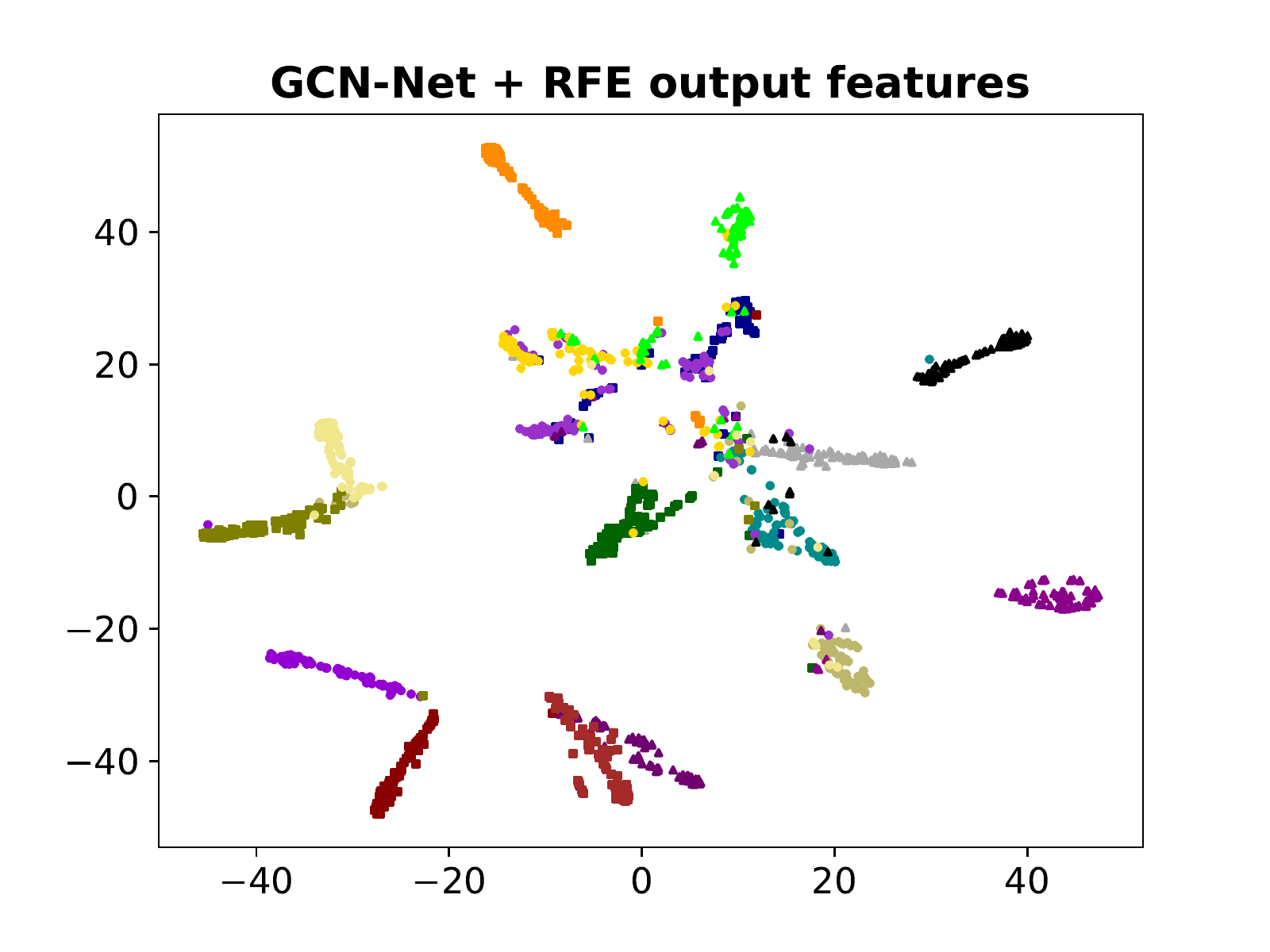} \\
    (a) \textbf{CNN-ResNet} & (b) \textbf{GCN-Net} & (c) \textbf{GCN-Net + RFE}  \\ 
    \end{tabular}
    \caption{Feature space illustrations computed by TSNE on Flower dataset with CNN-ResNet descriptor. It shows the (a) original feature space, (b) feature space obtained with the GCN, and (c) feature space obtained by the GCN using the RFE (our proposed approach) embeddings.}
    \label{fig:tsne_flowers}
\end{figure*}

In addition to the numerical analyses, qualitative experiments are also important for understanding the results achieved by the proposed approach. 
For better visualization of the improvements provided by RFE in the semi-supervised classification experiments, Figure~\ref{fig:tsne_flowers} illustrates feature spaces on Flowers17 dataset with CNN-ResNet descriptor for three different cases: \emph{(a)} features extracted by the CNN-ResNet descriptor; \emph{(b)} GCN-Net output features after being trained on the CNN-ResNet features; and  \emph{(c)} GCN-Net output features after being trained on the CNN-ResNet features combined to the RFE embeddings.
The TSNE method was used in order to compute the coordinates in the 2D space.
While each dot represents a different element of the dataset, each combination of color and shape corresponds to a distinct class.
Notice that \emph{(c)} presents the best correspondence among the visual groups formed by the dots and the original dataset classes.
This evinces our hypothesis that the RFE embeddings improve the classification of GCNs.

\begin{figure*}[!th]
    \begin{minipage}{\linewidth}
    \centering
    \includegraphics[width=.8\textwidth]{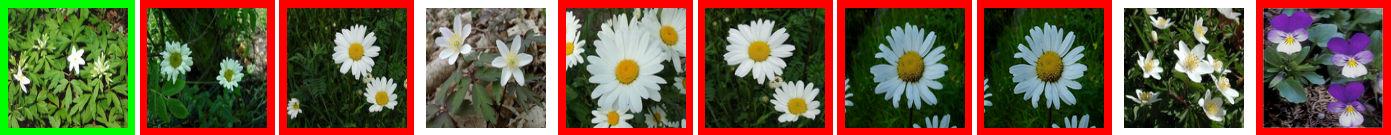} \\
    \includegraphics[width=.8\textwidth]{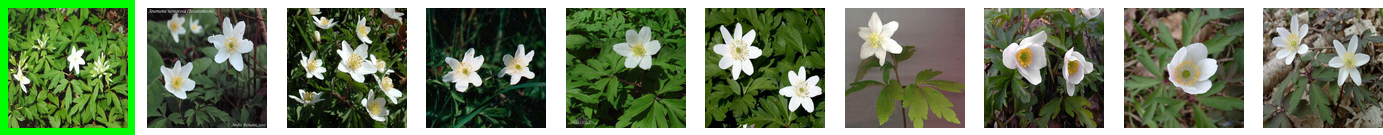} \\
    (a) Flowers Dataset \\
    \vspace{0.25cm}
    \includegraphics[width=.8\textwidth]{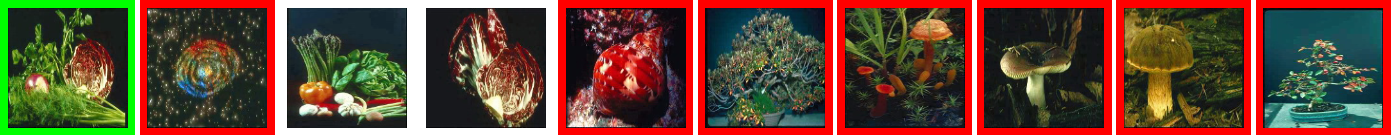} \\
    \includegraphics[width=.8\textwidth]{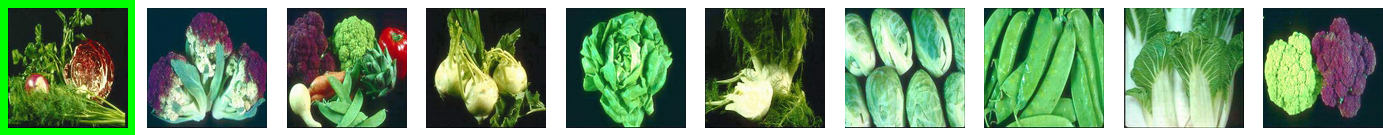} \\
    (b) Corel5k Dataset \\
    \vspace{0.25cm}
    \includegraphics[width=.8\textwidth]{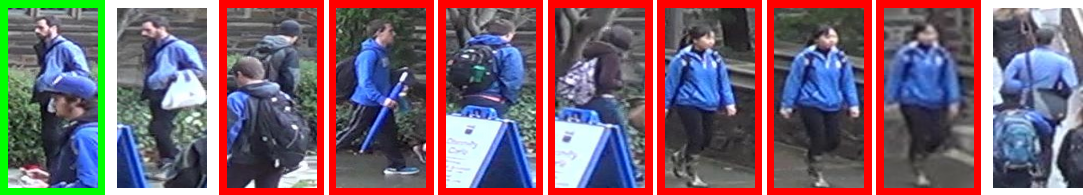} \\
    \includegraphics[width=.8\textwidth]{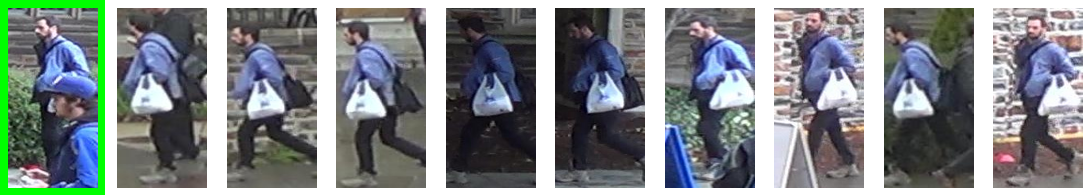} \\
    (c) Duke Re-ID Dataset \\
    \end{minipage}
    \caption{Examples of ranked lists before and after RFE was applied for three datasets. Query images are highlighted with green borders and wrong results are with red borders.}
    \label{fig:visual_rk_examples}
\end{figure*}

Experiments were also conducted to visualize the performance of RFE in retrieval tasks.
Figure~\ref{fig:visual_rk_examples} presents examples of ranked lists before and after the execution of our proposed method. These results were obtained on different datasets (CNN-ResNet for Flowers and Corel5k; and OSNET-AIN for Duke) with the default parameters and $k$. The query images are presented with green borders and the incorrect ones with red borders. It clearly shows the significant improvements for all the queries.

%%%%%%%%%%%%%%%%%%%

%-----------------------------------------------------
\section{Conclusion}
\label{sec:conclusion}
%-----------------------------------------------------

In this work, we have proposed Rank Flow Embedding (RFE).
The method is based on different techniques (hypergraph, Cartesian product, connected components) and can be used for improving both retrieval and classification tasks.
An extensive experimental evaluation was conducted on 10 datasets, including 7 general purpose and 3 person re-identification datasets.
The results are very promising for the vast majority of cases when compared to the state-of-the-art.
In future work, we intend to investigate new strategies for graph modeling and embedding generation.
We also intend to apply our method to other types of multimedia data.

% or


%\appendices

%\section{}

%\section*{Acknowledgment}

%The authors are grateful to São Paulo Research Foundation - FAPESP (grants \#2017/25908-6, \#2018/15597-6, and \#2020/11366-0), Brazilian National Council for Scientific and Technological Development - CNPq (grant \#309439/2020-5), Petrobras (grant \#2017/00285-6), and Microsoft Research for financial support.
%We also acknowledge the NSF Grant No. IIS-1814745.

\ifCLASSOPTIONcaptionsoff
  \newpage
\fi




%***************************************** BIBLIOGRAPHY ************************************************
%\small
%\def\baselinestretch{.9}
%\bibliographystyle{elsarticle-num} 
\bibliographystyle{IEEEtran}
\bibliography{references}

%\bibliography{reid_bib,EE,EarlyFusion,BibLegacy,BibReckNNGraphCC,BibHypergraph,CNN}
%*******************************************************************************************************

\end{document}